\documentclass[10pt, fleqn, twocolumn]{article}

\usepackage[hang,flushmargin]{footmisc} 
\usepackage[utf8]{inputenc}
\DeclareUnicodeCharacter{2212}{$-$}
\usepackage{graphicx}
\usepackage[fleqn]{amsmath}
\usepackage[version=4]{mhchem}
\usepackage{siunitx}
\usepackage{longtable,tabularx}
\usepackage{nccmath}
\usepackage{amsmath}
\usepackage{array}
\usepackage{booktabs}
\usepackage[superscript]{cite}
\usepackage[margin=1in]{geometry}
\usepackage[toc,page]{appendix}
\usepackage{setspace}
\usepackage{titlesec}
\usepackage{amssymb}
\usepackage{tabu}
\usepackage{listings}
\usepackage{afterpage}
\usepackage{titlesec}
\usepackage[labelfont=bf]{caption}
\usepackage{fancyhdr}
\usepackage{pgf}
\usepackage{float}
\usepackage{subcaption}
\usepackage{cleveref}
\usepackage{textcomp}
\usepackage{pict2e}
\usepackage{wasysym}
\usepackage{multicol}
\usepackage{lipsum}
\usepackage{indentfirst}
\usepackage{makecell} 

\usepackage[justification=centering]{caption}
\titleformat{\section}
{\normalfont\normalsize\bfseries}{\thesection}{0.5em}{}
\titleformat{\subsection}
{\normalfont\normalsize\bfseries\em}{\thesubsection}{0.5em}{}
\titleformat{\subsubsection}
{\normalfont\normalsize\bfseries\em}{\thesubsubsection}{0.5em}{}
\titlespacing*{\section}
{0pt}{3mm}{2mm}
\titlespacing*{\subsection}
{0pt}{2mm}{2mm}

\begin{document}

\twocolumn[
  \begin{@twocolumnfalse}

\begin{flushright}
\Large

\end{flushright}
\begin{centering}      
\large 

\vspace{4mm}

\textbf{Accelerating Deep Learning Applications in Space}\\
\vspace{0.5cm}
\normalsize 

Martina Lofqvist, Jos\'e Cano\\
School of Computing Science, University of Glasgow\\
18 Lilybank Gardens, Glasgow, United Kingdom; +44 (0)1413301640\\
2265215L@student.gla.ac.uk\\

\vspace{0.5cm}
\centerline{\textbf{ABSTRACT}}
\vspace{0.3cm}
\end{centering}

Computing at the edge offers intriguing possibilities for the development of autonomy and artificial intelligence. The advancements in autonomous technologies and the resurgence of computer vision have led to a rise in demand for fast and reliable deep learning applications. In recent years, the industry has introduced devices with impressive processing power to perform various object detection tasks. However, with real-time detection, devices are constrained in memory, computational capacity, and power, which may compromise the overall performance. This could be solved either by optimizing the object detector or modifying the images. In this paper, we investigate the performance of CNN-based object detectors on constrained devices when applying different image compression techniques. We examine the capabilities of a NVIDIA Jetson Nano; a low-power, high-performance computer, with an integrated GPU, small enough to fit on-board a CubeSat. We take a closer look at the Single Shot MultiBox Detector (SSD) and Region-based Fully Convolutional Network (R-FCN) that are pre-trained on DOTA – a Large Scale Dataset for Object Detection in Aerial Images. The performance is measured in terms of inference time, memory consumption, and accuracy. By applying image compression techniques, we are able to optimize performance. The two techniques applied, lossless compression and image scaling, improves speed and memory consumption with no or little change in accuracy. The image scaling technique achieves a 100\% runnable dataset and we suggest combining both techniques in order to optimize the speed/memory/accuracy trade-off.\\

\\
  \end{@twocolumnfalse}
]

\section*{INTRODUCTION}

Over the past decade, we have seen a dramatic decrease in the cost of accessing space and a proliferation of the number of satellites in orbit. The introduction of nanosatellites, a small satellite with a wet mass between 1 to 10 kg, has increased the research and development in this field (Buchen and DePasquale, 2014). As a result, more satellites are now orbiting Earth and they are downlinking terabytes of data each day. This raw data is then captured by ground stations located around the world; an expensive process considering that not all of the data is useful. Furthermore, the downlinking of data can be unreliable, restricted by the data rate, and geographically limited to the locations of ground station networks. By developing space applications for on-orbit processing of raw data, we can minimize this bottleneck to save both time and money. Furthermore, autonomous space systems offer many opportunities to continue exploring our solar system and beyond. Space exploration provides social, economic, and intellectual contributions to humanity. With the recent announcements from space agencies and companies around the world to revisit the Moon and prepare for manned missions to Mars\noindent{\footnote{\noindent https://www.nasa.gov/feature/nasa-unveils-sustainable-campaign-to-return-to-moon-on-to-mars}}, research related to artificial intelligence and the development of deep learning applications in space have received increased attention. 

Deep learning applications in space can have many use cases, such as the ability to self-navigate for collision avoidance or satellite maintenance, surface observation and modeling, automated satellite docking, and asteroid mining. These applications play a vital role for the future of space exploration and offer many opportunities including image analysis, classification, clustering and active learning (Mcgovern and Wagstaff, 2011). This paper focuses on computer vision, specifically object detection from satellite imagery. Object detection deals with detecting instances of a certain class in images. It has received increased attention in the past years due to the importance of its task (Tomayko, 1988), but also because of the notable advances in this area with the introduction of deep convolutional neural networks (DCNNs) and the arrival of powerful accelerators such as GPUs. 

However, object detection from satellite imagery remains a challenging task due to the variable sizes of the objects in the images and complex backgrounds. Furthermore, performing these tasks in space on resource constrained devices provides limitations, specifically in terms of memory and compute capacity. Therefore, our objective is to investigate ways to improve the performance in terms of speed and memory consumption. One approach would be to optimize the object detection model or other layers of the so-called Deep Learning Inference Stack (Turner, 2018). Another approach would be to modify the satellite data through image compression techniques. 

\subsection*{Motivation}
Space is becoming more accessible and as a result, more companies and research institutions are investigating applications for deep learning to space data and developing space-ready devices to place in orbit. Overall, deep learning applications performed on edge computing devices in space remains largely unexplored and only a few attempts to unveil its opportunities have been tested in orbit. Therefore, there are a lot of opportunities to perform further research in this area. Specifically, there has been little research conducted on neural networks on low-powered GPUs, and no research, as far as we know, on how to compress satellite data to optimize performance in the context of constrained devices.

\subsection*{Objectives}
The objective of this research is to evaluate different image compression techniques applied to satellite data and to find the trade-off between speed, memory consumption, and accuracy when running these images through object detectors on constrained devices. The aim is to reduce the inference time and memory consumption without a significant loss in accuracy. We will work with a device small enough to fit on board a CubeSat, a standardized miniature satellite which comes in a 10x10x10cm configuration, and find a large, annotated dataset consisting of satellite imagery that is pre-trained on object detection models. We will establish a baseline by running the images on powerful GPUs to evaluate the performance against. The research aims to explore the advantages of computing at the edge in order to expedite the development of intelligence in space.

\subsection*{Paper Structure}

\begin{itemize}
    \item The next section provides background information on the different topics that will be explored and it discusses related research that has been conducted in this area. Specifically, it gives an overview of deep learning and edge computing as well as present relevant research on these topics in the space sector. This provides insights on areas to explore further and motivate this research. We investigate running large satellite images through object detection networks on constrained devices. 
    \item The following section provides an overview of the project and discuss the problem in detail.
    
    \item The Design and Implementation section proposes a solution with different techniques for approaching the problem. The overall aim is to find ways to optimize deep learning applications on constrained devices. We design and implement methods for analyzing image compression techniques on different object detection networks. A baseline will be established to compare the results against.
    
    \item In the Evaluation section, performance is assessed in terms of speed, accuracy, and memory consumption of the applied techniques and results are discussed. 
    
    \item Finally, the Conclusion provides a summary of the research, reflects on the approach to the problem, and suggests future work in this field.
\end{itemize}

\section*{BACKGROUND \& RELATED WORK}

\subsection*{Deep Learning}
\begin{figure}[t]
    \centering
    \includegraphics[width=0.95\columnwidth]{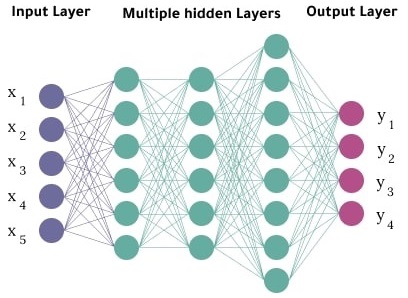}
    \caption{Deep Neural Network with multiple hidden layers.}
    \label{fig:deepnet}
\end{figure}

Artificial Intelligence (AI) powers many aspects of modern society. It is a technique that enables computers to mimic human behavior. Machine learning is a subset of AI that uses statistical methods for improving the machine by experience. One technique for implementing machine learning models is known as deep learning. Deep learning algorithms permit the machine learning application to train itself to perform tasks by exposing multilayered neural networks to large amounts of data. Neural networks are inspired by the structures of animal brains and comprise of ’neurons’ that combines multiple inputs to produce an output (Goodfellow et al., 2019). Generally, these output values are then fed as input to other neurons in a layered architecture. All neural networks consist of an input layer, output layer, and zero or more hidden layers. The difference between a neural network and a deep neural network is that the latter has multiple hidden layers, as illustrated in Figure \ref{fig:deepnet}. Each hidden layer computes the weighted sum of inputs through an activation function and transforms the input to produce an output. Hidden layers are set up in different ways to generate various results. Convolutional Neural Networks (CNNs) use convolutional layers to detect patterns, for example in images, by applying different filters (also called kernels). Some widely used deep learning frameworks are  TensorFlow\noindent{\footnote{https://www.tensorflow.org/}}, Keras\noindent{\footnote{https://keras.io/}}, PyTorch\noindent{\footnote{https://pytorch.org/}}, MXNET\noindent{\footnote{https://mxnet.apache.org/}}, Caffe\noindent{\footnote{https://caffe.berkeleyvision.org/}}, Theano\noindent{\footnote{http://deeplearning.net/software/theano/}}, and CNTK\noindent{\footnote{https://docs.microsoft.com/en-us/cognitive-toolkit/}}. 
    
\subsection*{Computer Vision}
A common application for deep learning is computer vision. The major computer vision techniques that help extract and analyze information include image classification, object detection, object tracking, semantic segmentation, and instance segmentation. In this paper, we focus on object detection. It is different from image classification in the sense that it detects multiple objects in an image and identifies their locations. The target location of the objects are usually presented by rectangular bounding boxes. In satellite images, the objects typically have various orientations and, therefore, some techniques output oriented bounding boxes (OBB) over the objects rather than horizontal bounding boxes (HBB). Training and evaluation require the dataset to contain both the images and the corresponding labels (the bounding box coordinates and the associated class category).

\subsection*{GPUs}
Running CNNs is a computationally expensive process. The networks can be accelerated by using a graphical processing unit (GPU). These devices are good at handling specialized computations that can run in parallel. By dividing a task into independent smaller computations that can be carried out simultaneously, a GPU is perfectly suited for neural networks. NVIDIA\noindent{\footnote{https://www.nvidia.com}} and AMD\noindent{\footnote{https://www.amd.com}} are two companies that develop GPUs. NVIDIA introduced the first GPU, GeForce 256, in the late 90’s. Since then, GPUs have become cheaper, more powerful, and smaller. They are now one of the core components of deep learning. NVIDIA has developed multiple embedded systems products that integrate GPUs for autonomous solutions\noindent{\footnote{https://www.nvidia.com/en-us/autonomous-machines/embedded-systems/}}. The NVIDIA Jetson platform powers a range of AI applications for computing at the edge. The platform is compatible with many of the major deep learning frameworks and NVIDIA provides tools, such as cuDNN\noindent{\footnote{https://developer.nvidia.com/cudnn}} and TensorRT\noindent{\footnote{https://developer.nvidia.com/tensorrt}}, to use them. Edge computing brings the computation and storage of data closer to the device where the data is being gathered. The advantages to edge computing systems is that they provide reduced latency and increased bandwidth, but at the expense of capacity and dependability.

\subsection*{Remote Sensing}
Remote sensing is the practice of making observations at a significant distance (Campbell and Wynne, 2011). In the space industry, this applies to meteorological, extraterrestrial, and Earth remote sensing. Observations from space platforms can help provide useful information and scientists are calling for the space industry to provide more data at a faster rate (Durrieu and R.F., 2013). The most common and accessible form is Earth observation, which can offer a lot of opportunities ranging from various forms of surface monitoring to disaster management. Remote sensing of the environment can be traced back to the early 1970s when Kondratyev et al. (1973) performed research on satellite images from Landsat 1. These remote sensing applications include the monitoring and management of agriculture, water, forests, fisheries, and the climate. The sensors collect data from a surface area either actively, by emitting energy, or passively, by gathering radiation emitted or reflected. It then observes the spectral differences in the data. Active sensors include LiDAR, GPS and RADAR, while passive sensors are typically photographic.

\subsection*{Computers in Space}

Computers have been used on spacecrafts since the middle of the 20th century. Initially introduced by NASA, the development of computing capabilities in space have served an important role in enabling a variety of crewed and unmanned missions, and have made a significant impact on improving overall computing technology (Tomayko, 1988). Over time, computers have been optimized in terms of size, weight, and power consumption, all of which are important features to take into consideration during the development of a mission.

\subsection*{Space-ready GPUs}

\begin{figure}[t]
    \centering
    \includegraphics[width=0.75\columnwidth]{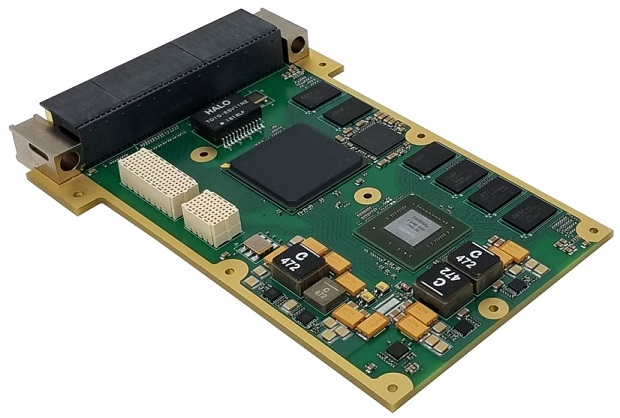}
    \caption{The EDGE GPU-based payload processor by IBEOS}
    \label{fig:ibeosgpu}
\end{figure}

In the space industry, GPUs have mainly been tested on the surface level. NASA uses NVIDIA GPUs for astronaut training in combination with VR techniques (Greenstein, 2016) as well as for running simulations in preparation for future human missions to Mars (Salian, 2019). Some laptops on the International Space Station (ISS) have an integrated GPU for Extravehicular Activity (EVA) preparation. In addition, SpaceX used the NVIDIA Tegra processor for powering touch screens on the Dragon V2 capsule to control the ship (Caulfield, 2014). The processing environment for spacecraft computers is radically different than on Earth (Tomayko, 1988). Putting computers in space is challenging due to radiation threats from solar flares or cosmic rays that have the power to ’bit flip’ and can trigger glitches. Thus far, GPUs have been proton tested, although further testing is required to achieve better results (Wyrwas, 2017). The company Ibeos\noindent{\footnote{https://ibeos.com/}} has developed a radiation-hardened GPU, EDGE\noindent{\footnote{https://ibeos.com/documents/cubesat/ Ibeos\_CubeSat\_GPU.pdf}}, that is small enough to fit inside a 1U or 3U CubeSat, that is yet to be tested in space (see Figure \ref{fig:ibeosgpu}).

\subsection*{Space Robotics}
A number of intelligent robots have been brought to and tested on the ISS, including the free flying robots SPHERES, a personal satellite assistant, Int-Ball, a robotics system developed by JAXA, CIMON, a social robot developed by ESA, and Astrobee, a robot system able to perform Intra Vehicular Activities (IVA) (Bualat et al., 2018). The Robonaut, delivered to the ISS in 2011 by NASA and General Motors, is capable of dexterous manipulation with the future aim of exceeding hand movements of suited astronauts (Bryndin 2019). Russia’s FEDOR Skybot F-850 is a multi-functional robot capable of performing basic tasks and interacting with the crew-members aboard the ISS (Bryndin, 2019). It has a predefined programme with elements of artificial intelligence, but is able to be switched to fully controlled mode if necessary. Machine learning is used to improve mobility of these robots to support IVAs and EVAs (Bryndin, 2019). Currently, their capabilities are limited as to mitigate any potential damages to the ISS.

\subsection*{Orbital Edge Computing}
To perform real-time processing on-board a nanosatellite, the device must be small yet powerful. The majority of research in deep learning applications on low power GPUs investigate different hardware and software choices for improving performance on these constrained devices (Loukadakis et al., 2018; Rovder et al., 2019; Gibson, 2019; Radu et al., 2019). Bradley and Brandon (2019) proposed a system for implementing on orbit processing. The Orbital Edge Computer (OEC) system replaces the traditional bent-pipe architecture of sending satellite data to Earth. Furthermore, it eliminates the bottleneck of requiring a significant number of ground-stations and downlinking large amounts of irrelevant data. This system includes an NVIDIA Jetson TX2, a small-sized board with a GPU that can fit inside a 1U CubeSat, and provides insight on software considerations.

\subsection*{Deep Learning in Space}
In regards to deep learning applications within the space industry, most have been performed on Earth’s surface using existing datasets or in preparation to test in orbit. An example of such is Evers’ (2019) work on satellite docking through object detection. In January 2019, researchers at the Tokyo Institute of Technology sent a deep learning attitude sensor to LEO and have plans on commercializing this type of technology in the future (Titech, 2018). ESA has been working on Mars rovers using AI and ML to navigate around obstacles autonomously and an AI assistant is being developed for the ISS by the German Aerospace Center (ESA, 2020). A common application for deep learning in the space industry is Earth observation. The number of these applications has surged over the past years, especially as the availability of and the quality of data has improved significantly. The quality of satellite images is measured by spatial resolution, the size of a single pixel on the ground. The level of detail depends on the spatial resolution and is measured by Ground Sample Distance (GSD), influenced by the camera focal length, pixel sensor size, and orbital altitude. Today, companies are capturing satellite images with a spatial resolution of less than one meter\noindent{\footnote{https://www.planet.com/products/planet-imagery/}}.

\subsection*{Object Detection}
Object detection in satellite images is a complex task due to the inconsistencies in the background of the images, the low object-to-pixels ratios, the orientations of the objects to be detected, and the low image resolutions. There is a wealth of open source satellite imagery, however, not a lot of the data is annotated. The limited amount of publicly available satellite datasets that are annotated makes training these object detection networks a challenging task. Existing datasets include:

\begin{itemize}
    \item TAS (Heitz and Koller, 2008)
    \item SZTAKI-INRIA (Benedek et al., 2012) 
    \item UCAS-AOD (Zhu et al., 2015) 
    \item HRSC 2016 (Liu and Mattyus, 2015)
    \item COWC (Mundhenk et al., 2016) 
    \item VEDAI (Razakarivony and Jurie, 2016) 
    \item 3K Vehicle Detection (Liu, Wang, Weng and Yang, 2016) 
    \item NWPUVHR-10 (Cheng et al. 2016) 
    \item xView (Lam et al., 2018) 
    \item DOTA (Xia et al., 2018)
    \item SpaceNET (Weir et al., 2019) 
\end{itemize}

Many of these datasets are only provided in ideal conditions and lack an adequate number of images, instances per image, and object categories. Therefore, most of the methods developed for object detection in Earth Vision are based on transfer learning and fine-tuning networks pre-trained on large datasets (Xia et al. 2018). Two datasets stand out in terms of size, number of annotations, and class categories: xView and A Large-Scale Dataset for Object Detection in Aerial Images (DOTA). xView has a very high resolution of 0.3 meters per pixel whilst DOTA has a variety of resolutions ranging from 0.2 to 1 GSD because of the images being collected from different sensors and satellite missions. Both were developed independently around the same time and offer contests in computer vision. 

The xView challenge\noindent{\footnote{http://xviewdataset.org/}} took place in 2018 and was held by the Defense Innovation Unit Experimental (DIUx)\noindent{\footnote{https://www.diu.mil/}} with the focus on national security and disaster response. Their work has about 40 citations on ArXiv\noindent{\footnote{https://arxiv.org/abs/1802.07856}} and about 10 of them have the term `Object Detection' in the title. The disadvantage to the xView dataset is that the number of objects are not equally distributed across the classes providing an extreme class imbalance. DOTA have almost 160 citations on ArXiv\noindent{\footnote{https://arxiv.org/abs/1711.10398}}, where about 60 of them have the term ’Object Detection’ in the title. Their contests\noindent{\footnote{https://captain-whu.github.io/DOTA/index.html}} have run for two consecutive years and their tasks relate to locating ground objects with either HBB or OBB. The most relevant research includes the work by Rotich et al. (2018) on image resizing and image splitting on the xView dataset. They propose a framework for implementing lightweight CNNs to detect and classify objects in high-resolution images. However, the research is conducted on a CPU and only provides insights on the mAP scores. It does not measure performance in terms of speed and memory consumption. 

The previous background research suggests a lot of opportunities for furthering the research on deep learning applications in space.

\section*{PROBLEM OVERVIEW}

\subsection*{Constrained Devices}
In the space environment, everything becomes more complex. Computers can show unexpected behaviors if the internal structures are changed due to the impacts of radiation. Furthermore, deep learning applications are typically run on the cloud with powerful GPUs that speed up the processing. However, in the space environment, devices are limited in terms of power and considering the vast increase in cost of sending a bigger and heavier satellite to space, the size of the edge computers matters significantly. Therefore, to perform on-orbit processing of data, the devices must be carefully considered. There exists some off-the-shelf low-powered GPUs, small enough in size to fit on a nanosatellite. The low power constraint provides another challenge as the device may be limited in the number and size of operations it is able to perform. This means longer execution times for processing the images. 

\subsection*{Large Satellite Data}
Satellite data can be extremely large. The pixel dimension of the images can be up to 8,000 pixels in height by 8,000 pixels in width and they can have a high spatial resolution. The quality of satellite images is continuously improving to extract more useful data, increasing the need for large-scale on board storage systems. Running these large images through a network on board the satellite can take up a lot of the available RAM. In some cases it could cause the device to run out of memory, which could lead to mission failure or loss. Furthermore, the processing time of these large images could be very long. A way to handle the problem with memory consumption and processing time is to optimize the neural network or to modify the data. However, when optimizing for speed and memory, it may be at the expense of accuracy. Applying image optimization techniques will change the features that can be extracted from the images. This is particularly important to note for small objects with a tiny pixel dimension.

\subsection*{Lack of Data and Biased Datasets}
The research on deep learning applications in space is limited and many of the neural networks are trained on natural images. This means that the object detection models are formed through transfer learning and fine-tuning of the networks. Therefore, finding a large dataset that contains annotated satellite images is challenging. In addition, the image data from optical sensors can vary greatly. Most available datasets consist of images with the same pixel dimensions. Networks that are trained on a specific input size could be biased. Finding a pre-trained model that is trained on images of different sizes is very challenging and to our knowledge, all networks that were found have been trained on satellite images of the same size. Thus, the input size of the network needs to be taken into consideration as research by Xia et al. (2018) showed that all images were cropped to the same size before training. Furthermore, no pre-trained network was found that had been tested on a low-powered GPU. Object detection in satellite images is very different from detection in natural scenes. This is due to the inconsistencies in the background of the images, the low object-to-pixels ratios, the orientations of the objects to be detected, and the low image resolutions. Many existing datasets are provided in ideal conditions and lack an adequate number of images, instances per image, and object categories. The dataset needs to be annotated and contain labels for evaluating the network. The object instances can be annotated differently, either as HBB or OBB.

\subsection*{Unfavorable Results}
\begin{figure}[t]
    \centering
    \includegraphics[width=0.95\columnwidth]{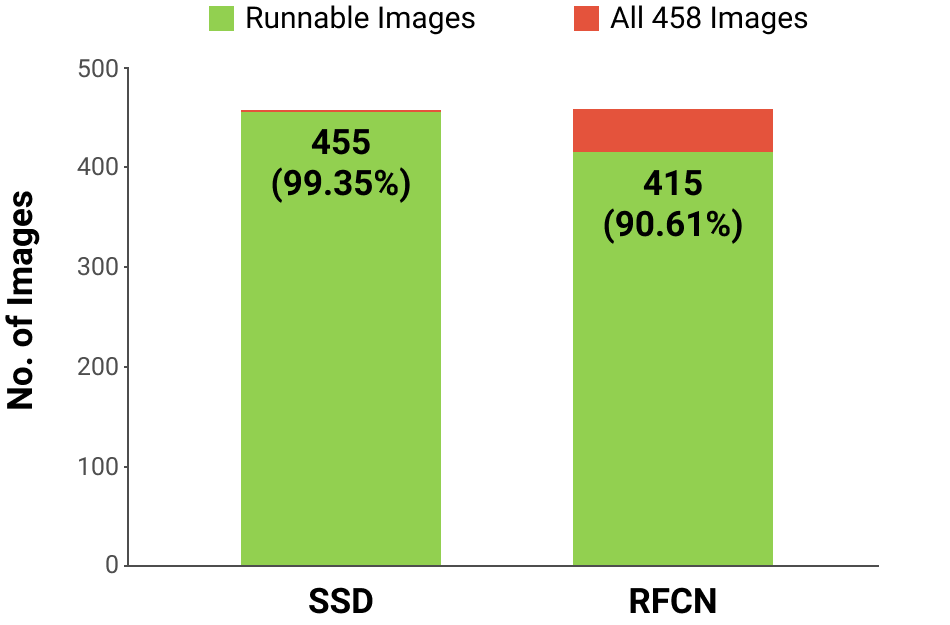}
    \caption{Runnable images of the DOTA dataset on the pre-trained SSD and R-FCN models. The figure shows that 99.35\% of the images can run on the SSD model and 90.61\% of the images can run on the R-FCN model}
    \label{fig:runnable}
\end{figure}

In order to run the object detectors on satellite images in a real-time environment, we need to reduce the demand of resources, including both the memory consumption and the execution time. When we run satellite images from the DOTA dataset on two pre-trained models, Single Shot Detector (SSD) and Region-based Fully Convolutional Network (R-FCN), on an NVIDIA Jetson Nano, we obtain unfavorable results. Our initial results show that the large images take a long time to process, sometimes up to 7.5 seconds per image, and consume a lot of memory, sometimes causing the device to produce an Out of Memory (OOM) error. About 10\% of the images in the dataset on the R-FCN network are entirely unable to be executed on the device due to their large sizes (see Figure \ref{fig:runnable}). We want to be able to run the full dataset as well as improve the processing speed and reduce the memory consumption.

\section*{DESIGN AND IMPLEMENTATION}
In this section we discuss the design choices and implementations made to approach the problem of running large images on constrained devices. We describe the compression techniques applied to the images to reduce the execution time and memory consumption. We chose to work with the DOTA and used the pre-trained object detectors SSD and R-FCN. The parallel computing will be performed on an NVIDIA Jetson Nano on the Deep Learning framework TensorFlow. The results will help establish the speed/ accuracy/memory consumption trade-off.

\subsection*{Device: Jetson Nano}
\begin{figure}
    \centering
    \includegraphics[width = 0.75\columnwidth]{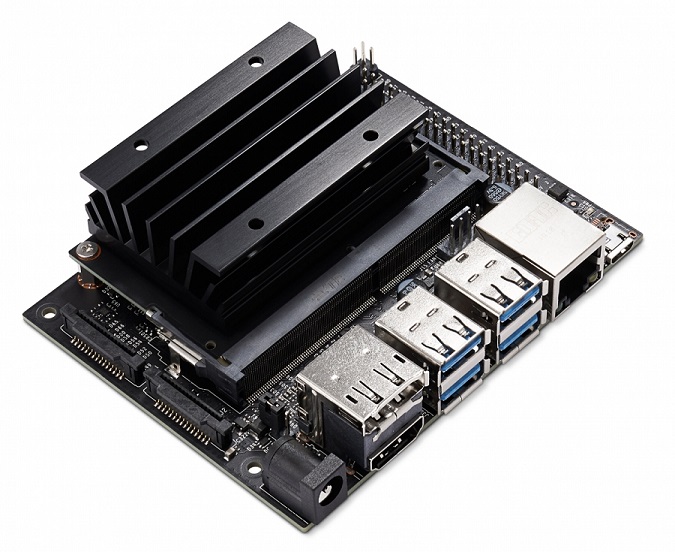}
    \caption{NVIDIA Jetson Nano}
    \label{fig:jetsonnano}
\end{figure}

We examine the capabilities of a NVIDIA Jetson Nano\noindent{\footnote{https://www.nvidia.com/en-us/autonomous-machines/embedded-systems/jetson-nano/}} (Figure \ref{fig:jetsonnano}); a low-power, high capability, efficient GPU, small enough to fit on-board a CubeSat (Venturini, 2017). This device comes with a 128-core integrated NVIDIA Maxwell GPU and a quad-core 64-bit ARM CPU. It has 4GB of LPDDR4 memory at 25.6GB/s and 5W/10W power modes. The device is configured by downloading and installing the Jetson Nano Developer Kit SD card image\noindent{\footnote{https://developer.nvidia.com/embedded/jetpack}}.

We flash an SD card with JetPack 4.2 and install the relevant packages in the same versions as described in the GitHub repository\noindent{\footnote{https://github.com/ringringyi/DOTA\_models}} that contains the pre-trained object detection models we chose to work with. The device is pre-configured with 4GB of SWAP. We keep the SWAP at 4GB since the device runs out of memory (OOM) when executing the object detection code on 0GB of SWAP for all images in the dataset. The device is then set up with SSH for remote access and Jupyter Notebook is used to run the object detection code.

\subsection*{Dataset: DOTA}
\begin{figure}[t]
    \centering
    \includegraphics[width = \columnwidth]{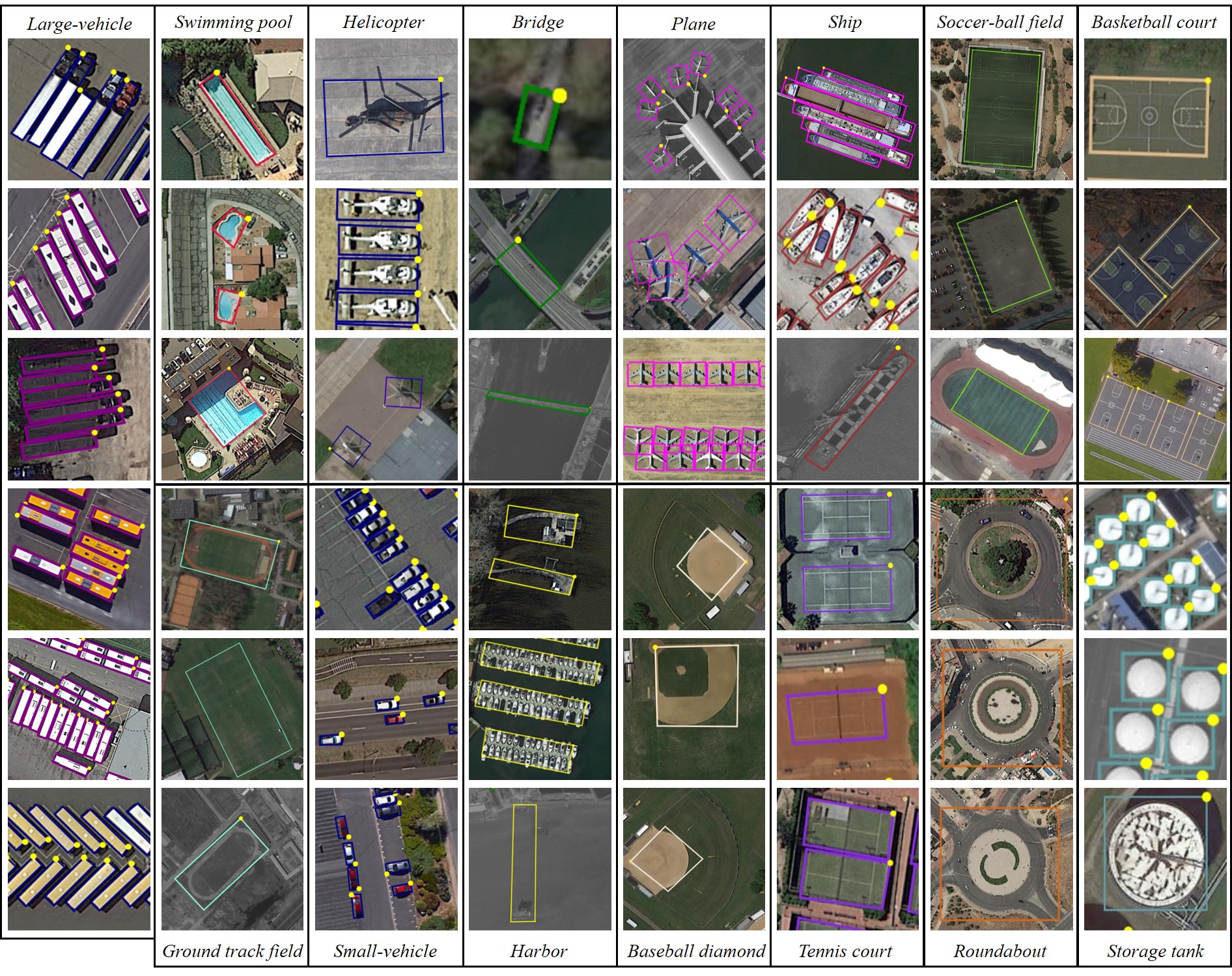}
    \caption{Samples of annotated instances and object categories in DOTA}
    \label{fig:dotainstances}
\end{figure}

We chose to work with DOTA as it is one of the most robust datasets, containing satellite images from different sensors, multiple annotations, and realistic settings (Xia et al. 2018). They also provide models that are trained on the dataset with state-of-the-art object detection algorithms. The DOTA-v1.0 dataset consists of 2,806 images with a total of 188,282 annotated instances from 15 class categories, as seen in Figure \ref{fig:dotainstances}. 

DOTA contains images with dimensions ranging from about $350\times 350$ to $7,500\times 7,500$ pixels and with natural scenes that makes this dataset relevant to real-world applications. The images in DOTA have a very high spatial resolution, capable of detecting small vehicles. The images are collected from different sensors and platforms to eliminate biases and are taken from a variety of locations to improve data diversity. The dataset is split into training (50\%), validation (33.3\%), and testing (16.7\%) sets.

\begin{figure}[h]
\centering
\begin{minipage}[b]{0.23\textwidth}
    \includegraphics[width=\textwidth]{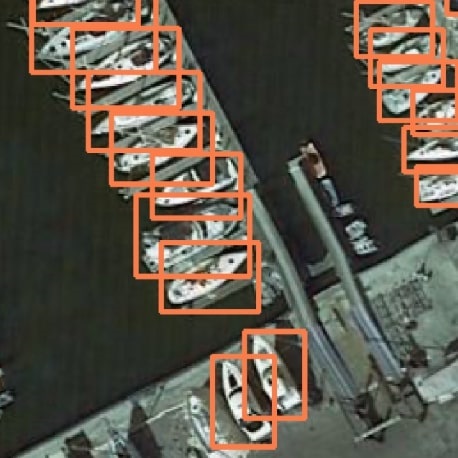}
    \label{fig:hbbships}
\end{minipage}
\begin{minipage}[b]{0.23\textwidth}
    \includegraphics[width=\textwidth]{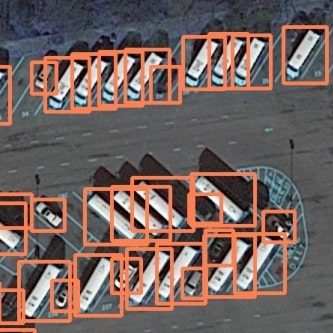}
    \label{fig:hbbvehicles}
\end{minipage}
\caption{Samples of crowded instances in annotated images in DOTA using horizontal bounding boxes (HBB)}
\label{fig:crowded-instances}
\end{figure}

Both the training set and validation set have ground truth labels for each image, the testing set does not. The instances are labelled both with horizontal bounding boxes (HBB for short) and oriented bounding boxes (OBB for short). The labels are in the following format: x1, y1, x2, y2, x3, y3, x4, y4, category, difficulty, where (x1, y1) is a set of coordinates representing one of the bounding box corners. The frequencies of instances vary across different image sizes where some small images may contain a lot of annotated objects whilst other larger images may only contain a handful. The average bounding box quantity is 67.10 per image. Certain objects, such as cars and ships, are particularly difficult to detect as they often appear in high density areas. This makes it challenging to draw bounding boxes around those objects (see Figure \ref{fig:crowded-instances}). In contrast, large objects that do not appear in crowded areas, such as swimming pools and tennis courts, are often easier to detect. In this project, we use the validation set which consists of 458 images and we use the labels for HBB, since that is what the publicly available models are trained on. The images are in PNG format and represented as 24-bit true color (RGB) images. The input layer size of the SSD is $608 \times 608$ pixels and of R-FCN it is $1,024 \times 1,024$ pixels. Since the images in the dataset represent a vast variety of sizes and there are very few squared images, this will have to be accounted for in the evaluation.

\subsection*{Models: SSD \& R-FCN}
The pre-trained models for DOTA include the following with source codes available on GitHub:     

\begin{itemize}
    \item \textbf{Object Detection Benchmarks for Aerial Images}\noindent{\footnote{https://github.com/dingjiansw101/AerialDetection}}
    \item \textbf{ROI Transfer}\noindent{\footnote{https://github.com/dingjiansw101/RoITransformer\_DOTA}}
    \item \textbf{Faster R-CNN OBB}\noindent{\footnote{https://github.com/jessemelpolio/Faster\_RCNN\_for\_DOTA}} with ResNet-101 backbone architecture
    \item \textbf{SSD}\noindent{\footnote{https://github.com/ringringyi/DOTA\_models}} with InceptionV2 backbone architecture
    \item \textbf{R-FCN}\noindent{\footnote{https://github.com/ringringyi/DOTA\_models}} with ResNet-101 backbone architecture
    \item \textbf{YOLOv2}\noindent{\footnote{https://github.com/ringringyi/DOTA\_YOLOv2}} with a customized GoogLeNet backbone architecture
\end{itemize}

They all use different frameworks. In our project, we chose to take a closer look at the pre-trained models for Region-based Fully Convolutional Network (R-FCN) and Single Shot MultiBox Detector (SSD) trained on HBB. This is because they are both trained on TensorFlow\noindent{\footnote{https://www.tensorflow.org/}}, which is compatible with the Jetson Nano board.

\begin{figure}[t]
    \centering
    \includegraphics[width=0.95\columnwidth]{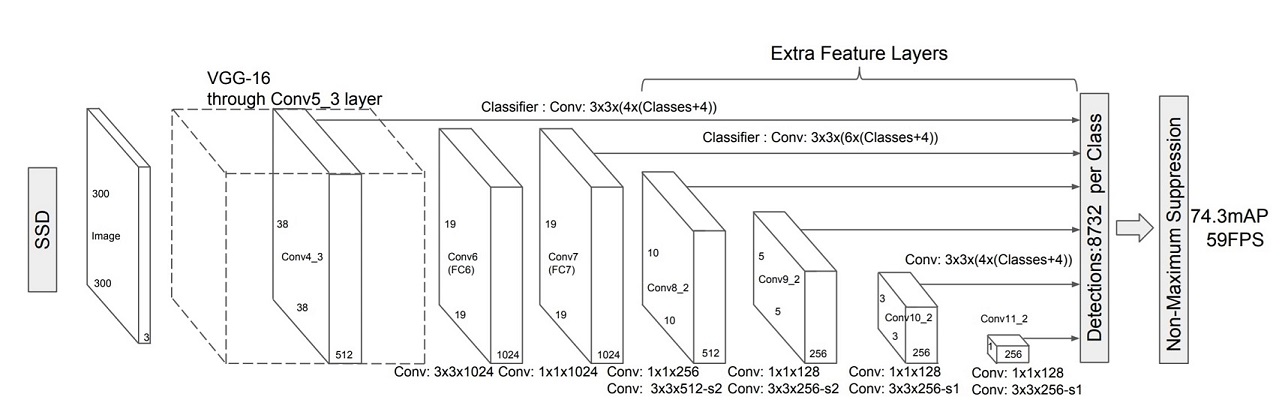}
    \caption{SSD Network Architecture}
    \label{fig:ssd}
\end{figure}

\textbf{SSD} predicts bounding boxes and confidence scores from a single pass (Liu, Anguelov, Erhan, Szegedy, Fu and Berg, 2016). It is a multibox detector with a single deep neural network (see Figure \ref{fig:ssd}). This means that SSD is able to predict objects of various scales by combining different feature maps and default boundary boxes. Feature maps at a higher resolution are responsible for finding smaller objects and therefore SSD performs better on images with high resolution. However, compared to other detection methods, the SSD network generally performs worse on small objects (Zhao et al., 2019).

\begin{figure}[t]
    \centering
    \includegraphics[width=0.95\columnwidth]{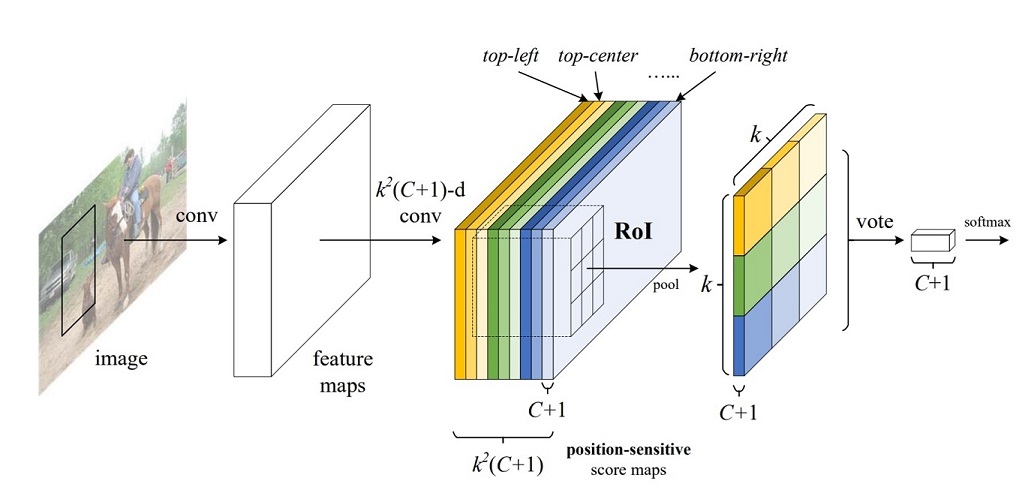}
    \caption{R-FCN Network Architecture}
    \label{fig:rfcn}
\end{figure}

\textbf{R-FCN} crops features from the last layer, prior to prediction, which means that per region computation is minimized which optimizes the speed. The cropping mechanism is location sensitive which improves the confidence score (Dai et al., 2016). Since the backbone architecture is the ResNet-101 model, R-FCN has 100 convolutional layers that are used to compute the feature maps, as seen in Figure \ref{fig:rfcn} (Dai et al., 2016).

\subsection*{Deep Learning Framework: TensorFlow} 
TensorFlow (Abadi et al., 2015) belongs to the group of most widely used deep learning frameworks. It is an open-source library with a flexible design that can run on the NVIDIA GPU through optimized kernels. The NVIDIA CUDA architecture provides the parallel computing platform for the execution of these kernels.

\subsection*{TensorFlow Object Detection}
SSD and R-FCN were trained using the publicly available TensorFlow Object Detection repository\noindent{\footnote{https://github.com/tensorflow/models/tree/master/\\research/object\_detection}}. Since both models were trained in 2017, they used an older version of the TensorFlow Object Detection code. In order to be consistent, we use the same version as what the networks were trained on and make slight modifications to the code. In short, the code imports the model, including the config file and frozen detection graph (CKPT), and the DOTA label map to add the correct label for each box detected. The threshold is set to 0.01, meaning that we capture close to 100\% of all detected objects. The network predicts the object by providing the confidence score, class probability, and bounding box coordinates at the final layer of the networks. The results are saved to a file to be used later for evaluation. The procedure is as follows:

\begin{enumerate}
    \item Start up the system
    \item Remove all SWAP to ensure no additional SWAP has been allocated with \texttt{\$ sudo swapoff -a}
    \item Add the SWAP back with \texttt{\$ sudo swapon -a}
    \item Connect to the Jetson Nano via SSH
    \item Open the Jupyter Notebook \texttt{.ipynb} file where the object detection code is located
    \item Run the image, record the SWAP memory and save the output
    \item Clear the output of the kernel
    \item Restart the system by writing \texttt{\$ sudo reboot}
\end{enumerate}

The SWAP allocation is validated through the program \emph{htop}\noindent{\footnote{https://hisham.hm/htop/}}, as illustrated in Figure 9. The first image that runs through the object detection code always takes longer as the system is warming up. So, we always discard the first image and, therefore, use the smallest image in the set to warm-up the system. We run each image 5 times for the baseline and 3 times for the modified images. We record the speed and memory consumption and then calculate the averages.

\begin{figure}[t]
    \centering
    \includegraphics[width=0.95\columnwidth]{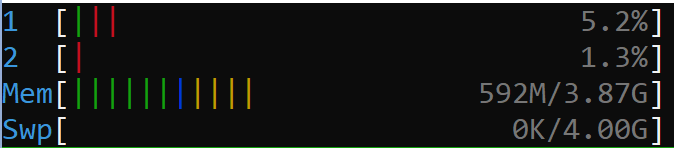}
    \caption{The \emph{htop} Program Displays Memory Usage, Including SWAP}
    \label{fig:htop}
\end{figure}

\subsection*{Image Manipulation}
Many of the images in DOTA are very large and therefore would take a long time to run on the CNNs and would consume a lot of memory on the constrained device that could lead to OOM errors. Previous work on this data crops the images into 608x608 or 1024x1024 patches (Xia et al., 2018). Cropping the images may not always be the best option as data may be lost if objects are cut into multiple parts and it could end up taking up more memory on the device. Therefore, we will instead be looking into different ways of modifying the images to decrease their size. These techniques are used to reduce the amount of data without compromising the quality of the image. 

\subsection*{Image Compression}
Compression techniques can be divided into two classes; lossy compression and lossless compression. These techniques are used for reducing the file size while pixel dimensions are kept the same. Some data is lost when images are compressed using the lossy compression method, while lossless compression can recover all data exactly to its original format. The mathematical basis of the two compression methods is information theory and different algorithms have been developed to implement them (Sayood, 2006). The algorithms encode the information in fewer bits than the original image, which makes the file size smaller. The unpredictable part, known as the prediction error, in this process is transmitted to the decoder that reconstructs the image based on a prediction model. This process is known as predictive coding. It captures the correlation amongst image pixels. The value of a pixel can be predicted based on some previous pixel values. An error image is given by subtracting the predicted pixel value from the current pixel value at the same spatial location. For lossless compression, the encoding process is error free. This can be achieved using techniques such as Huffman coding and arithmetic coding (Bawa, 2010). For lossy compression, the image is first quantized before the information is encoded. The level of compression can be predetermined and it is a trade-off between file size and the speed of encoding/decoding. 

\subsection*{Image Scaling}
A scaling technique will be applied to modify the pixel width and height of the image. The image ratios remain the same. This technique is also known as image interpolation, which works by using existing data to predict values at unknown points. It tries to achieve the best approximation by using information from surrounding pixels. There are several interpolation algorithms that can be grouped into adaptive algorithms and non-adaptive algorithms (Kim et al., 2009). The adaptive algorithms changes while the non-adaptive treats all pixels the same. Examples of adaptive algorithms include Bilinear and Nearest Neighbor (Kim et al., 2009). The Bilinear method provides better quality of the image, while the Nearest Neighbor is faster. The quality of the downsized image depends on the original image. Many details and a high resolution of the original image will not reduce the quality significantly.

\subsection*{Our Implementation}
We evaluate lossless image compression and the scaling technique to the images in the dataset. The largest image in the dataset is resized to a size that allows it to run successfully on both models. Other images will be resized and compressed to 30\%, 50\%, and 80\% of the original image size to evaluate the changes to the performance. The images will be resized using the Microsoft Photos App\noindent{\footnote{https://www.microsoft.com/en-us/p/microsoft-foton/\\9wzdncrfjbh4}}, which allows for free modification of the pixel dimensions (height and weight). The original images have a bit depth of 24 (True Color, RGB), however, the Microsoft Photos app saves the resized images with a depth color of 32 (True Color, RGBA, transparent). Since the object detection networks take an input of images with a 24-bit color, the images need to be converted. A free online image converting platform\noindent{\footnote{https://online-converting.com/image/}} proves to be useful for performing this task. The lossless compression technique is applied to the images using an online tool where specifying the compression level is possible and the PNG format is sustained.

\subsection*{Evaluation Methods}
\begin{figure}[t]
    \centering
    \includegraphics[width =0.75\columnwidth]{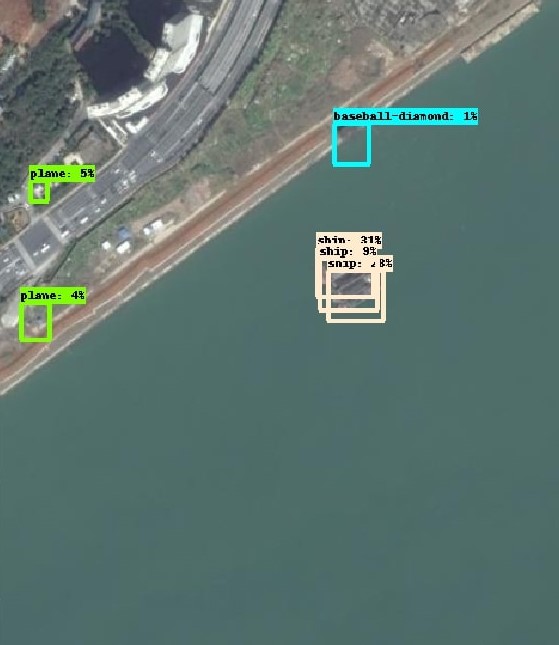}
    \caption{Instances Detected in Image P2310 after Running it through Model SSD. Each Instance has a Bounding Box with the Associated Class and Confidence Score}
    \label{fig:ssdobjects}
\end{figure}

The performance is evaluated according to memory consumption, speed, and accuracy. The memory consumption is observed by running the program \emph{htop} and observing the SWAP memory at the end of the object detection process. The results in terms of inference and information about the object detected in each detection are given as output after running the TensorFlow object detection code. The inference speed is recorded by adding a time function before and after the detection part of the code is executed. The detection output provides the confidence score, class category, bounding box coordinates, and the image with the detected bounding boxes, as illustrated in Figure \ref{fig:ssdobjects}. The confidence score is provided as a percentage in a range determined by the threshold set in the code (between 1\% and 100\% for a threshold of 0.01). Class categories are predetermined and provided by the \texttt{dota\_label\_map.pbtxt} file that maps an id to a name. The coordinates of each bounding box are calculated and returned in the form \texttt{(xmin, xmax, ymin, ymax)}. This information is then stored in \texttt{.txt} files to be retrieved by the evaluation scripts. The methods for evaluating the results include precision, recall, average precision (AP), and Intersection over Union. The code was developed to implement each of these techniques and the scripts were run on Colab\noindent{\footnote{https://colab.research.google.com/}}.

\subsection*{Precision and Recall}
Precision and recall are two common metrics to use in evaluating the performance of a given classification model. Typically, this is established after first creating a confusion matrix of True/False and Positive/Negative values. $TP$ stands for True Positive, $FP$ stands for False Positive, $TN$ stands for True Negative, and $FN$ stands for False Negative.

Precision calculates the positive predicted values given by the ratio of true positive values (TP) and the total value of positive predictions. The formula for precision is as follows:

\begin{equation}
    \quad \quad \quad \quad \quad \quad {Precision} = \frac{TP}{TP+FP}
\end{equation}

Recall calculates the true positive rate or sensitivity given by the ratio of true positive values (TP) and the total ground truths positives. The formula for recall is as follows:

\begin{equation}
    \quad \quad \quad \quad \quad \quad {Recall} = \frac{TP}{TP+FN}
\end{equation}

\subsection*{Mean Average Precision}
The mean average precision (mAP) is a method for evaluating the accuracy of a network and is calculated by finding the area under the curve. The average precision (AP) is the precision averaged across recall values between 0 and 1.0. The general definition for AP is as follows:

\begin{equation}
   \quad \quad \quad \quad \quad \quad {AP} = \frac{1}{11}\int_{0}^{1} p(r)dr,
\end{equation}

where $AP$ is the average precision, $p$ is the precision, $r$ is the recall, and $dr$ is set to 1 if the $r$th item has the same label as the previous one, otherwise it is set to 0.

Since precision and recall are always between 0 and 1, so is the AP score. There are two ways for calculating the AP; Interpolated AP and AP (Area Under Curve AUC). We use the first approach and implement a function for this. This is done by finding the precision value for 11 points from 0 to 1.0 on the recall axis and then calculating the average. The Mean Average Precision (mAP) is calculated by averaging across all categories. We make no distinction between AP and mAP since the AP will already be averaged over all categories.

\subsection*{Intersection over Union (IoU)}
\begin{figure}[t]
    \centering
    \includegraphics[width = 5cm]{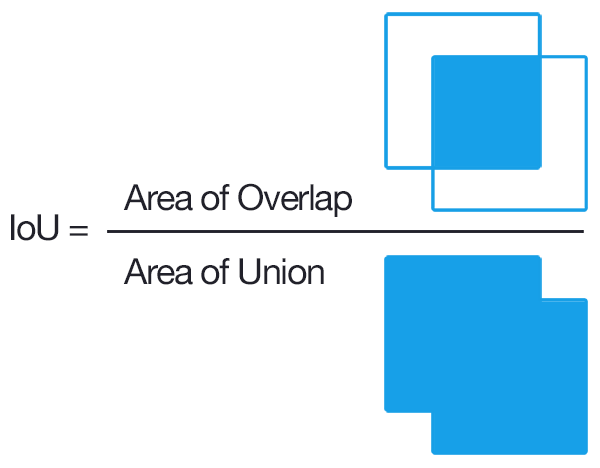}
    \caption{The IoU Calculation}
    \label{fig:iou}
\end{figure}

Typically, when calculating the AP for object detection, intersection over union (IOU) is used. This is the ratio of the area of intersection and area of overlap of a predicted bounding box and ground truth bounding box, as illustrated in Figure \ref{fig:iou}\noindent{\footnote{https://www.pyimagesearch.com/2016/11/07/\\intersection-over-union-iou-for-object-detection/}}. This is then used to determine whether the predicted bounding box is $TP$, $FP$, $TN$, or $FN$ according to the IoU threshold. The threshold represents the level of intersection. For example, an IoU threshold of 0.5 considers bounding boxes to be correct if the intersection is above 50\%. The prediction is classified as a TP when the IoU is above the threshold, an FP when IoU is below the threshold or when there are duplicate bounding boxes, a TN when there are no objects in the image, and an FN when IoU is above the threshold but has the wrong classification.

COCO\noindent{\footnote{http://cocodataset.org/\#detection-eval}} presents three methods for calculating the AP value using the IoU threshold:

\begin{itemize}
    \item AP: AP at IoU= 0.50: 0.05: 0.95 (IoU starts at 0.50 and increases to 0.95 with steps of 0.05)
    \item AP@IoU=0.50 (traditional way of calculating)
    \item AP@IoU=0.75 (IoU of BBs need to be $>$ 0.75)
\end{itemize}

\begin{figure}[t]
    \centering
    \includegraphics[width =0.75\columnwidth]{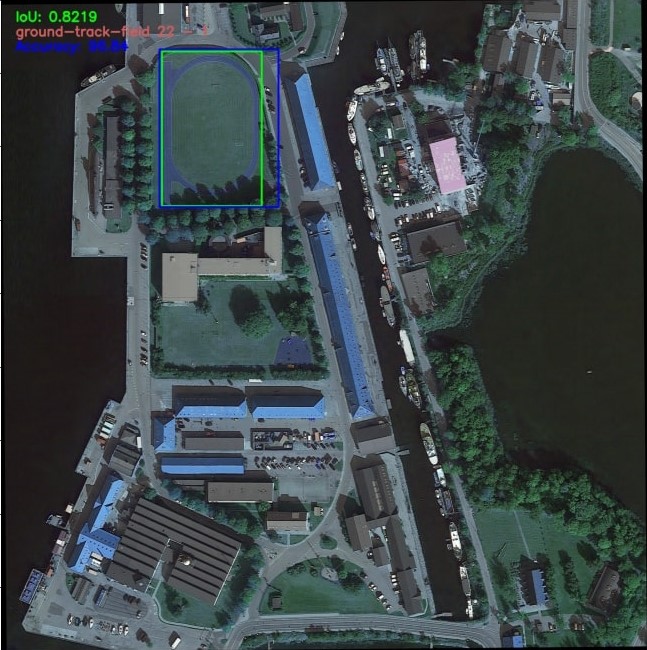}
    \caption{Image P2794 with the IoU output of 82.19 \% and Confidence Score of 96.84\% for the detected ‘ground-track’ class}
    \label{fig:p2794-all}
\end{figure}

There are also other ways of determining the IoU threshold. R-FCN uses a threshold of 0.3 IoU while SSD uses one of 0.5 IoU. We calculate the IoU at 0.1 to get all possible values and we then apply methods for calculating the mAP using different IoU thresholds. Furthermore, we also consider the confidence scores of the detected objects, often not considered by other evaluations. There are also classes in the code that display the image with the ground truth bounding box (in green) and predicted bounding box (in blue) as illustrated in Figure \ref{fig:p2794-all}. The image also displays the IoU (in green text), the object numbering (in red text) with the ground truth object number to the left and predicted to the right, and the confidence score (in blue text). The accuracy of the predictions (the number of objects accurately detected) are calculated by combining the IoU and confidence scores. Since there may be duplicate predictions for a given ground truth bounding box, we take the one with the highest product of IoU and confidence score.

\subsection*{Baseline System}
The performance of the Jetson Nano is compared against a baseline. This baseline measures the inference of the models by running them on a cluster containing two GPUs; an NVIDIA Titan RTX and an NVIDIA Titan V. Table \ref{tab:hardware} shows the technical specifications of the cluster and the Jetson Nano.

Firstly, the images are organized by total pixel size, since the memory consumed and inference time of running the images on the networks is influenced by the number of pixels in an image rather than the size of the file. This value is calculated by multiplying the pixel height and pixel width of each image. The reason for this is to find the largest images in the dataset and test whether they are able to run successfully on both models in order to determine what percentage of the dataset is runnable. By ‘successfully’ we mean that the images do not cause an Out of Memory (OOM) error. The largest image in the validation set is P1854 and the smallest image is P2310. Since all images are able to run successfully on the cluster, we begin by assessing the percentage of images that are capable of running on the Jetson Nano. Images that cause an Out of Memory (OOM) error on the Jetson Nano are recorded. The OOM error (also displayed as Resource Exhaust) means that both the RAM and the SWAP memory were consumed, which causes the Jupyter Notebook to shut down. In graphs to follow, images that cause an OOM error will be represented by a red ‘X’. 

We find that about 1\% of the images on the SSD network and 10\% of the images of the images on the R-FCN network are unable to run (see Figure \ref{fig:runnable}). The difference in the values between the two models is because R-FCN is a more resource consuming network due to its additional number of convolutional layers. The largest runnable image on the R-FCN network is P2794.

\begin{table}
\caption{Hardware Platforms}
\fontsize{7.5}{6.5}\selectfont
\begin{tabular}{|p{1.28cm}|p{1.75cm}|p{1.75cm}|p{1.3cm}|}
    \hline
     \textbf{} & \textbf{Titan RTX} & \textbf{Titan V} & \textbf{Jetson Nano}\\
     \hline
     \textbf{GPU Memory} & 24GB GDDR6 & 12GB HBM2 & 4GB LPDDR4\\
     \hline 
    \textbf{Cores} & \parbox[t]{5cm}{4608 (CUDA),\\ 576 (Tensor), \\72 (RT)} & \parbox[t]{5cm}{5120 (CUDA),\\ 640 (Tensor)} & 128 (CUDA)\\

     \hline 
     \textbf{Memory Bandwidth} & 672 GB/s & 652.8 GB/s & 25.6 GB/s\\ 
     \hline
     \textbf{TFLOPs} & 130 & 110 & 0.5\\
     \hline
\end{tabular}
\label{tab:hardware}
\end{table}

\begin{figure}[t]
    \centering
    \includegraphics[width=0.95\columnwidth]{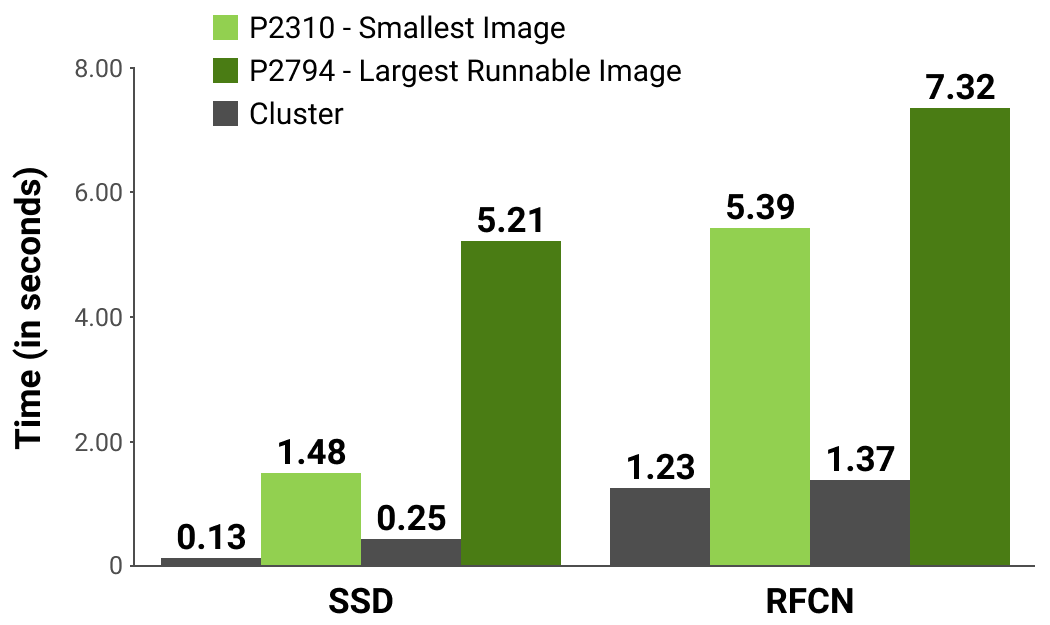}
    \caption{Average Speed (in seconds) of images P2310 and P2794, models SSD and R-FCN, compared to the baseline (cluster)}
    \label{fig:bas_inf}
\end{figure}

Next, we run the smallest image and largest runnable image on both the cluster and the Jetson Nano to compare the inference time. The results show that the speed is significantly compromised on the Jetson Nano for both images and models (see Figure \ref{fig:bas_inf}). The two images are on average 16x slower on the SSD model and 5x slower on the R-FCN model. These results are unfavorable.

\begin{figure}[t]
\centering
\begin{subfigure}{.43\columnwidth}
  \centering
  \includegraphics[width=\columnwidth]{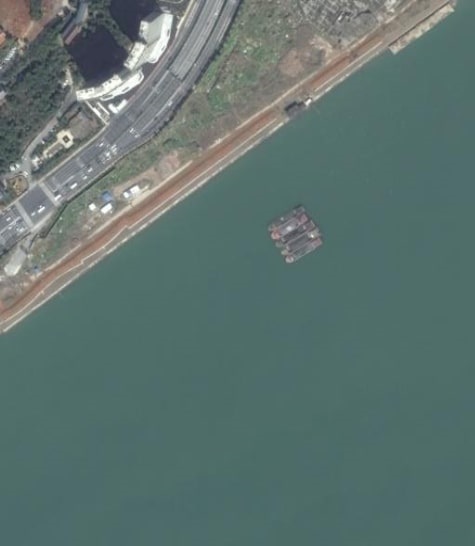}
  \subcaption{P2310}
  \label{fig:p2310}
\end{subfigure}
\hspace{1mm}
\begin{subfigure}{.49\columnwidth}
  \centering
  \includegraphics[width=\columnwidth]{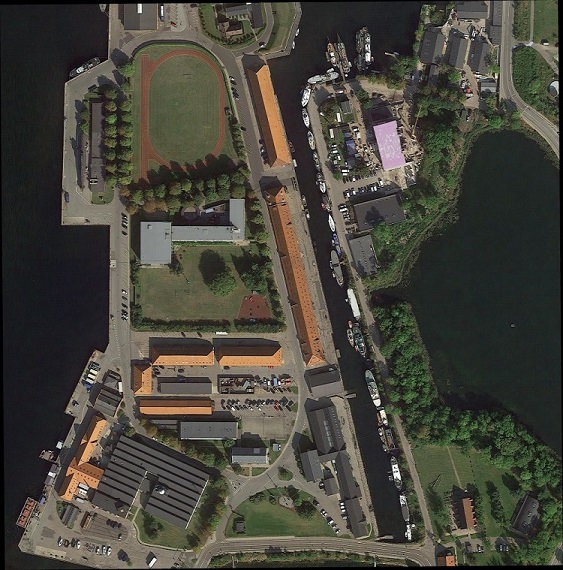}
  \subcaption{P2794}
  \label{fig:p2794}
\end{subfigure}
\newline
\begin{subfigure}{0.95\columnwidth}
  \centering
  \includegraphics[width=\columnwidth]{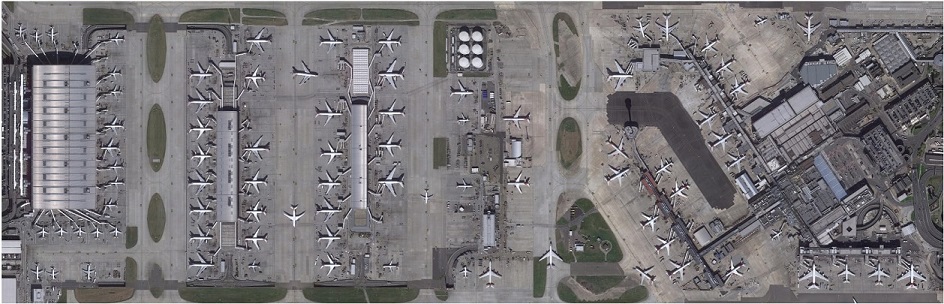}
  \subcaption{P1854}
  \label{fig:p1854}
\end{subfigure}
\caption{Images in the DOTA dataset where a) is the smallest image in the validation set, b) is the largest image capable of running on both models, and c) is the largest image in the validation set according to pixel size}
\label{fig:base_images}
\end{figure}

By applying different compression techniques to the images, the inference time is expected to improve. We also assess the impact on performance in terms of accuracy and memory consumption. The performance of the compressed images are compared against a baseline established by running the original images on the Jetson Nano before applying any compression technique. The following images will be assessed (see Figure \ref{fig:base_images}):

\begin{itemize}  
\item Smallest Image: \textbf{P2310} - 259,350 pixels ($475\times 546$) - size 209kB
\item Largest Image: \textbf{P1854} - 57,372,921 pixels ($13,383 \times 4,287$) - size 41.7MB
\item Largest Runnable Image: \textbf{P2794} - 19,504,872 pixels ($4,392 \times 4,441$) - size 97.8MB
\end{itemize}

\begin{figure}[t]
    \centering
    \includegraphics[width = \columnwidth]{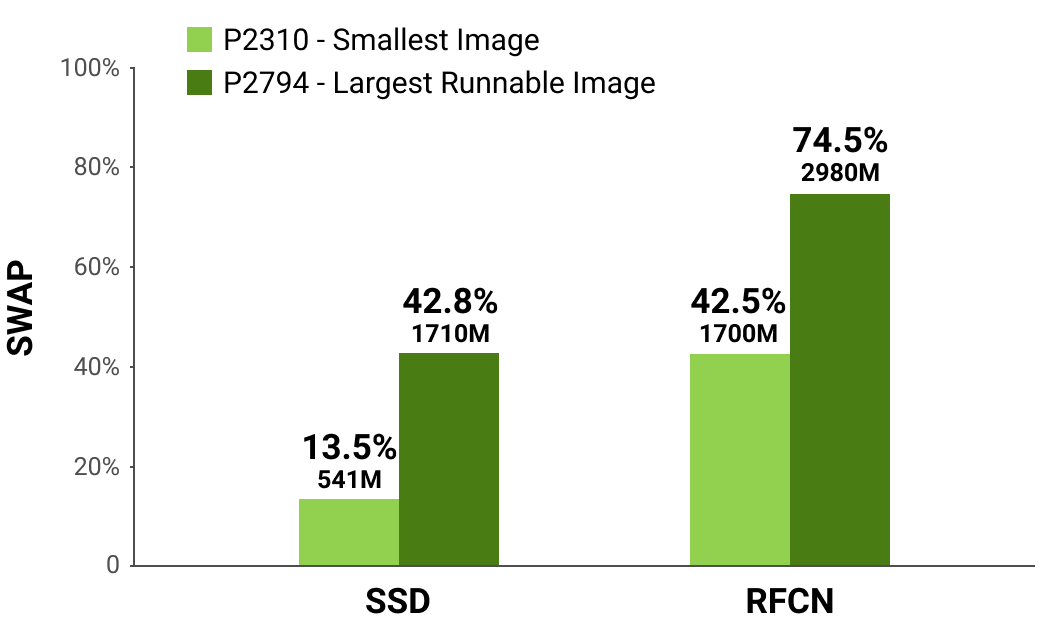}
    \caption{Memory Consumption (in Megabytes) of images P2310 and P2794, models SSD and R-FCN}
    \label{fig:bas_swap}
\end{figure}

\begin{figure}
    \centering
    \includegraphics[width =\columnwidth]{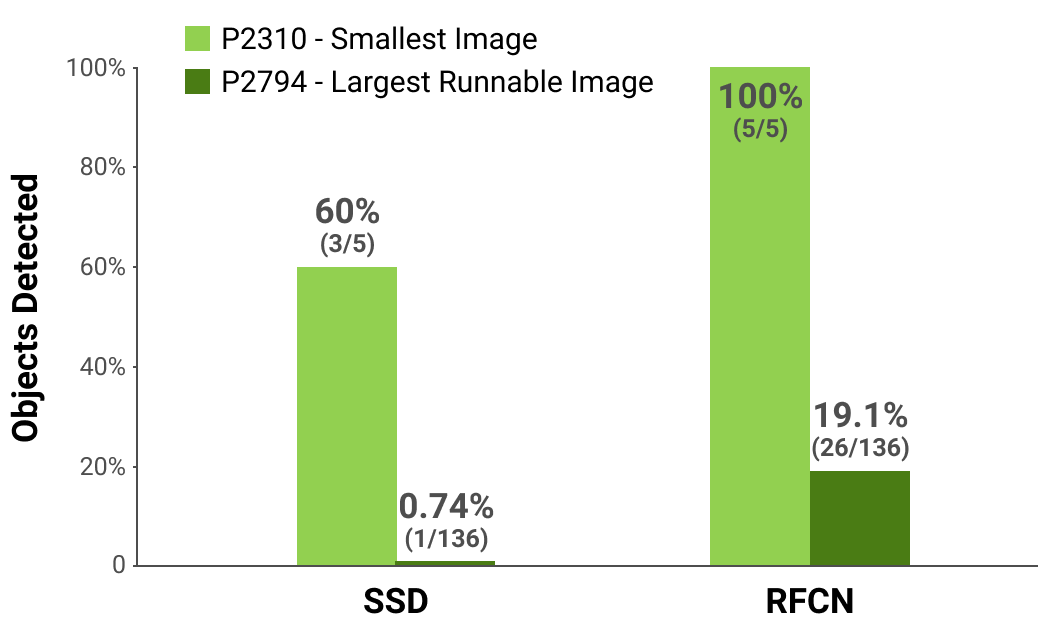}
    \caption{Average Accuracy of images P2310 and P2794, models SSD and R-FCN}
    \label{fig:bas_acc}
\end{figure}

The baseline results show that the smallest image, P2310, runs faster than the largest runnable image, P2794, on both networks, as expected (see Figure \ref{fig:bas_inf}). Image P2310 is 4 seconds (almost 4 times) faster than P2794 on the SSD network and about 2 seconds faster on the R-FCN network. The SSD model is faster than the R-FCN model for both images. This is due to the architectures of the networks, as discussed in an earlier section and will be further detailed below. The memory consumption shows a similar relationship between the images and models (see Figure \ref{fig:bas_swap}). Less memory was consumed by the smallest image compared to the largest image in both models as well as for SSD compared to R-FCN. Applying compression techniques to these images should decrease the file size and thus consume less memory. In terms of accuracy, the value was established by calculating the number of accurate objects detected with a confidence score of 50\% or more, using the IoU tool. Across all images, the R-FCN network showed higher accuracy than the SSD network (see Figure \ref{fig:bas_acc}).

Overall, the results show that the R-FCN network can solve more complex problems, although it is slower at detecting the objects and consumes more memory. This is explained by the differences in the network architectures. R-FCN is fully convolutional with multiple hidden layers, more than SSD. Furthermore, since SSD is a single stage detector, it means that it will typically perform worse in terms of accuracy compared to a multi-stage detector such as R-FCN (Wu and Li, 2019). In the results by Xia et al. (2018), we can see that the mAP of R-FCN (52.58) is about 2x higher than that of SSD (29.86). Therefore, higher accuracy of R-FCN and faster prediction of SSD are expected. Our baseline results on the Jetson Nano show that the mAP for R-FCN on the smallest image is 27.6 and for SSD is 23.3. R-FCN also has a higher mAP for the largest runnable image (24.0) compared to SSD (2.3). The average of the mAP over the two images for each model shows that the mAP for R-FCN (25.8) is twice as high as for SSD (12.8). This is consistent with the results presented by Xia et al. (2018).

The higher accuracy in image P2310 compared to image P2794 is associated to the differences in objects with respect to instance sizes. This is what makes predictions of objects in satellite images a challenging task. Most of the objects detected correctly in image P2794 on the R-FCN network were large and included object categories such as 'ground-track-field' and 'ships'. None of the 90 objects under the category 'small-vehicles' were detected. On the contrary, the object under the category 'ground-track-field' was accurately detected and 25 out of the 54 (46\%) objects under the category 'ships' were detected. The SSD network only detected the object under the ’ground-track-field’ category, which is about 15 times larger in pixel size than the second largest object that was accurately detected. For this reason, some networks are evaluated using different AP scores, according to small, medium, and large sized object categories. 

The differences in time and memory consumption between the smallest image and largest runnable image suggests that improvements can be achieved by modifying the images. We want to achieve this without compromising the accuracy. The first task is to successfully run the full dataset. This is done by determining at what percentage the largest image, P1854, needs to be compressed to in order for it to run successfully on both models. We then apply both compression techniques to retrieve three new images at 80\%, 50\%, and 30\% of the original image (see Table \ref{tab:specs}).

\begin{table}
\caption{Size Specifications for the Modified Images}
\fontsize{7.5}{6.5}\selectfont
\setlength{\aboverulesep}{0pt}
\setlength{\belowrulesep}{0pt}
\centering
\begin{tabular}{|c|c|c|}
    \toprule
     \textbf{} & \textbf{Resized Image} & \textbf{Compressed Image}\\
     \midrule
     \textbf{P2310 80\%} & 110 KB & 165 KB\\
     \midrule 
     \textbf{P2310 50\%} & 53.3 KB & 157 KB\\
     \midrule
     \textbf{P2310 30\%} & 23.6 KB & 154 KB\\ 
     \midrule
     \textbf{P2794 80\%} & 21.0 MB & 31.8 MB\\
     \midrule
     \textbf{P2794 50\%} & 9.18 MB & 31.7 MB\\
     \midrule
     \textbf{P2794 30\%} & 3.40 MB & 31.0 MB\\
     \midrule
     \textbf{P1854 80\%} & 49.1 MB & 70.0 MB\\
     \midrule
     \textbf{P1854 50\%} & 22.0 MB & 67.9 MB\\
     \midrule
     \textbf{P1854 30\%} & 8.57 MB & 64.9 MB\\
     \bottomrule
\end{tabular}
\label{tab:specs}
\end{table}

\section*{EVALUATION}

\subsection*{Run Full Dataset}
In order to run the full dataset, we resize the largest image, P1854, until it is capable of running on the slower R-FCN network. The results show that when the largest image is resized to 59\% of its original size, with a pixel dimension of $7896 \times 2529$, the image is able to run on both networks without any OOM errors. The average time for this resized image on the R-FCN network is 8.23 seconds, and 60 out of 221 objects (27\%) were detected accurately. The lossless compression technique was unable to compress this image to a small enough size for it to run successfully on both models. Applying a lossy compression technique would solve this issue, however, this technique is typically unfavorable as the images are unable to be restored to their original version.

\subsection*{Image Compression}
\begin{figure}[t]
\centering
\begin{subfigure}{.95\columnwidth}
  \centering
  \includegraphics[width=\columnwidth]{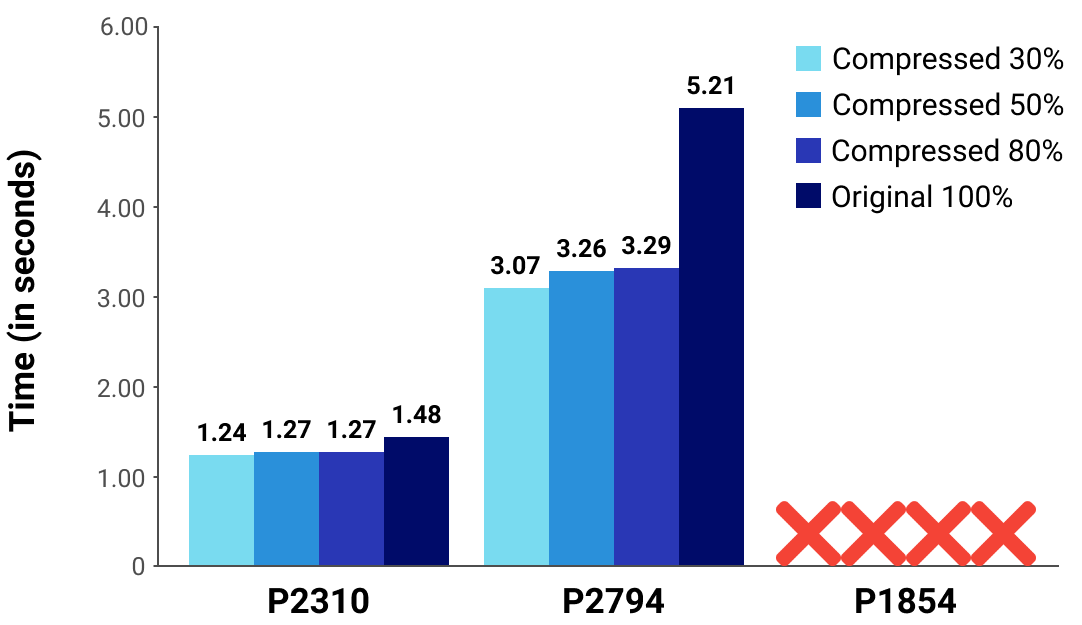}
  \subcaption{SSD}
  \label{fig:comp_speed_ssd}
\end{subfigure}
\newline
\begin{subfigure}{.95\columnwidth}
  \centering
  \includegraphics[width=\columnwidth]{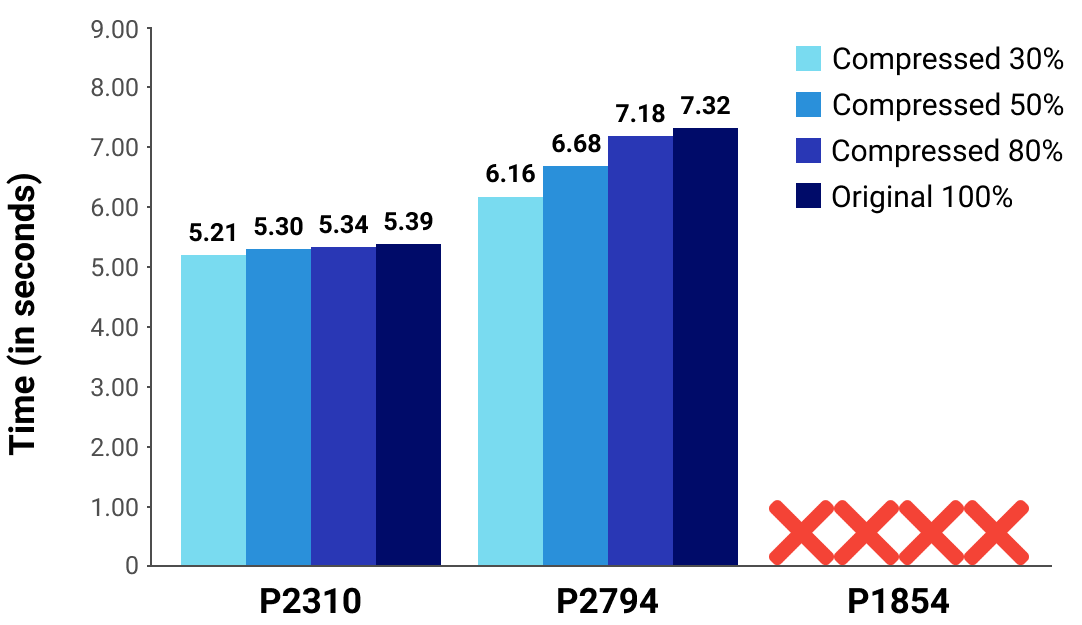}
  \subcaption{R-FCN}
  \label{fig:comp_speed_rfcn}
\end{subfigure}
\caption{Average Speed (in seconds) of the compressed images in both models. The ‘X’ means that the GPU ran out of memory (OOM). The images improve in speed when resized to a smaller version}
\label{fig:comp_speed}
\end{figure}

The results of compressing the images to 80\%, 50\%, and 30\% show that in the majority of cases, the speed was improved when the compression rate increased (see Figure \ref{fig:comp_speed}). Image P1854 was unable to run on either of the two models due to the remaining large size after the compression technique had been applied, leading the device to run out of memory (OOM). Image P2794 improved significantly in speed from the original image to the first compressed image for SSD. Image P2794 improved with 2.14 seconds for SSD and 1.16 seconds for R-FCN from the original image (100\%) to the most compressed image (30\%). Image P2310 improved with 0.24 seconds for SSD and 0.18 seconds for R-FCN from the original image (100\%) to the most compressed image (30\%).

\begin{figure}[t]
\centering
\begin{subfigure}{.95\columnwidth}
  \centering
  \includegraphics[width=\columnwidth]{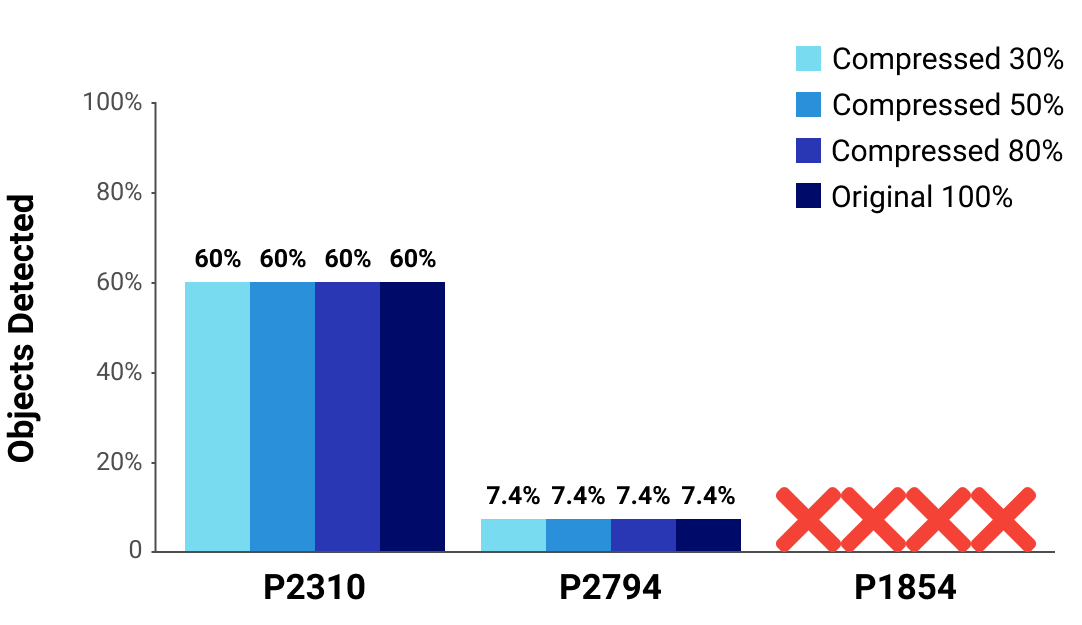}
  \subcaption{SSD}
   \label{fig:comp_acc_ssd}
\end{subfigure}
\newline
\begin{subfigure}{.95\columnwidth}
  \centering
  \includegraphics[width=\columnwidth]{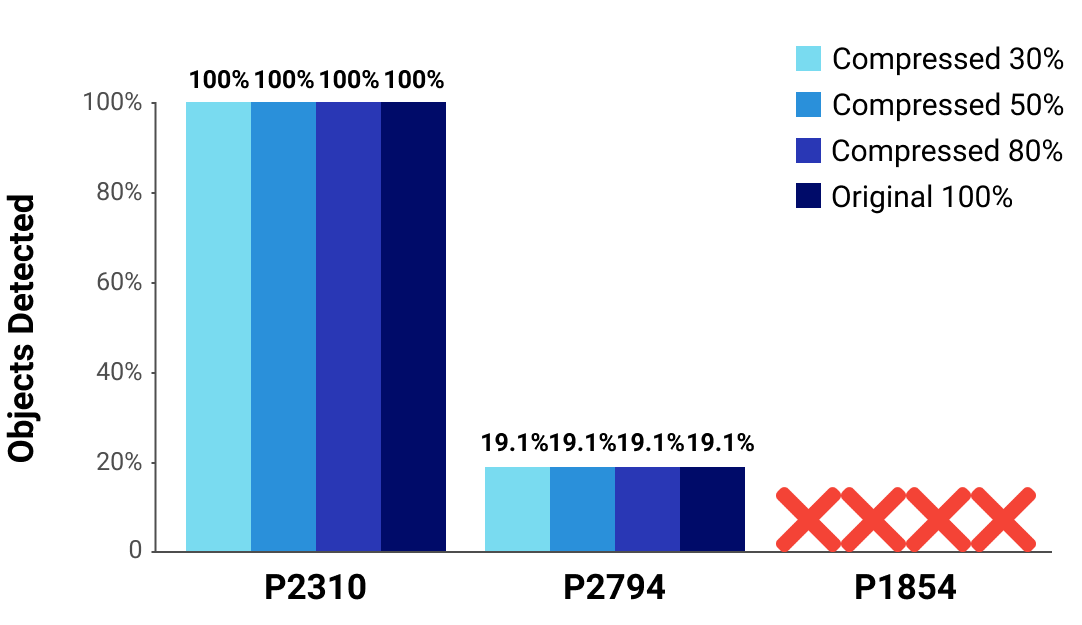}
  \subcaption{R-FCN}
  \label{fig:comp_acc_rfcn}
\end{subfigure}
\caption{Objects Accurately Detected in the compressed images in model SSD. The ‘X’ means that the GPU ran out of memory (OOM)}
\label{fig:comp_acc}
\end{figure}

In terms of accuracy, the performance was identical to that of the original image for the number of objects detected and their respective confidence scores (see Figure \ref{fig:comp_acc}). This would suggest that the accuracy correlates to the total pixel size. We test this on a resized image, P2794 resized to 50\%, to see if the accuracy differs when applying a lossless compression technique to this image. Results show that the two images do not differ in terms of accuracy, however, the inference time and memory consumption are reduced. This makes sense since the image size is less and it contains less data after the lossless compression technique has been applied.

\subsection*{Image Resize}
\begin{figure}[t]
\centering
\begin{subfigure}{.95\columnwidth}
  \centering
  \includegraphics[width=\columnwidth]{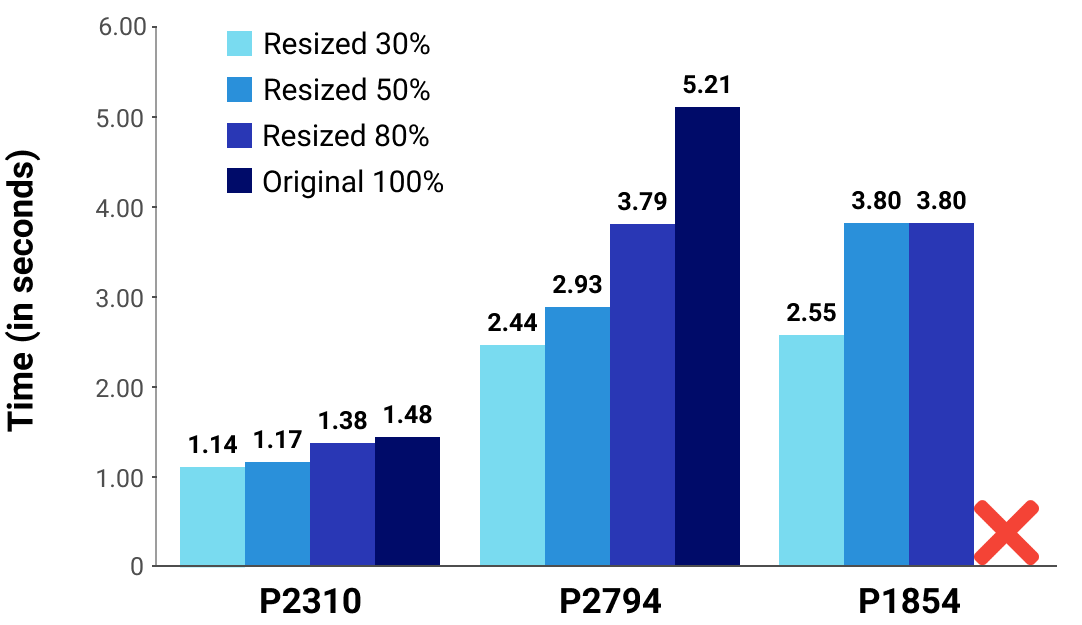}
  \subcaption{SSD}
   \label{fig:res_speed_ssd}
\end{subfigure}
\newline
\begin{subfigure}{.95\columnwidth}
  \centering
  \includegraphics[width=\columnwidth]{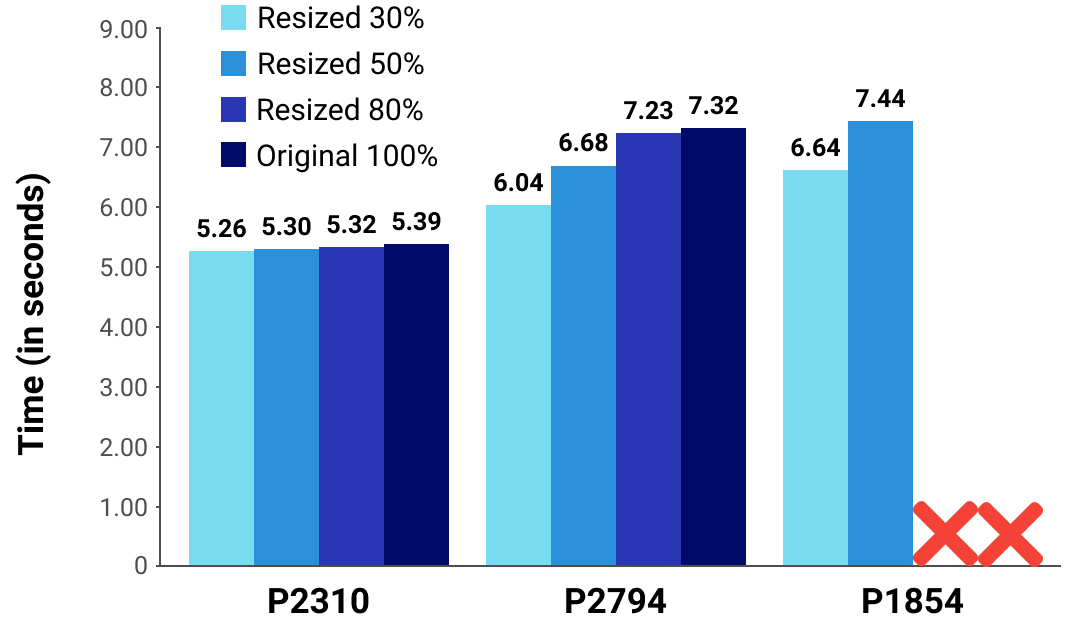}
  \subcaption{R-FCN}
  \label{fig:res_speed_rfcn}
\end{subfigure}
\caption{Average Speed (in seconds) of the resized images in both models. The ‘X’ means that the GPU
ran out of memory (OOM). The images improve in speed when resized to a smaller version}
\label{fig:res_speed}
\end{figure}

The results of resizing the images to 80\%, 50\%, and 30\% show that in the majority of cases, the speed was improved after applying this compression technique (see Figures \ref{fig:res_speed}a and \ref{fig:res_speed}b). Image P2794 improved with 2.77 seconds for SSD and 1.28 seconds for R-FCN from the original image (100\%) to the most compressed image (30\%). Image P2310 improved with 0.34 seconds for SSD and 0.13 seconds for R-FCN from the original image (100\%) to the most compressed image (30\%). The inference time for the compressed images did not match that of the baseline and further compression would be necessary in order to reach those results.

\begin{figure}[t]
\centering
\begin{subfigure}{.90\columnwidth}
  \centering
  \includegraphics[width=\columnwidth]{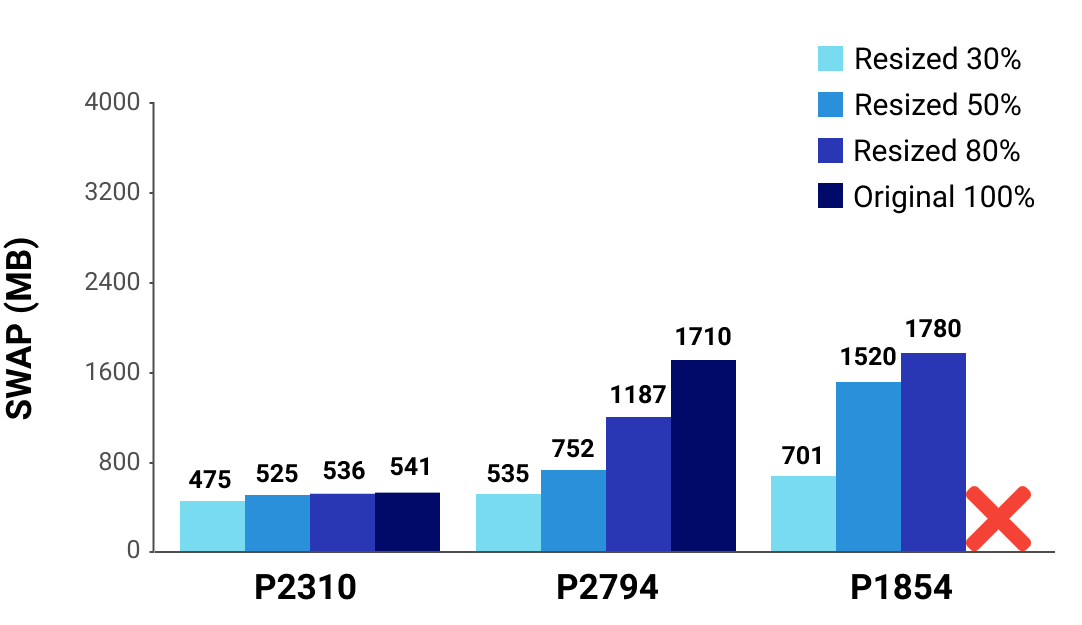}
  \subcaption{SSD}
   \label{fig:res_swap_ssd}
\end{subfigure}
\newline
\begin{subfigure}{.90\columnwidth}
  \centering
  \includegraphics[width=\columnwidth]{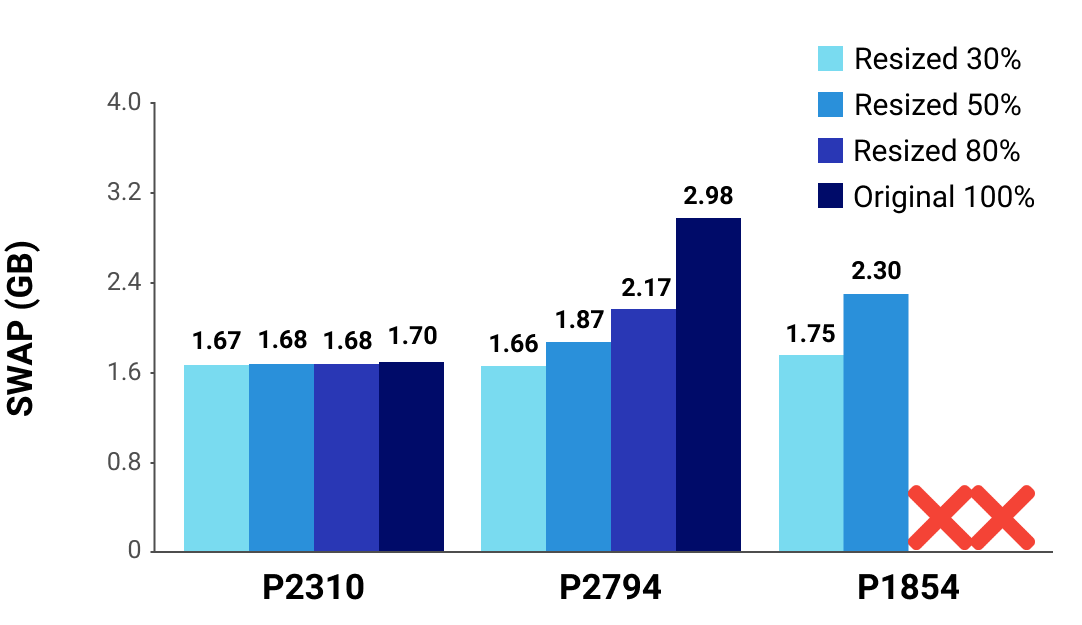}
  \subcaption{R-FCN}
  \label{fig:res_swap_rfcn}
\end{subfigure}
\caption{Average SWAP of the resized images in both models. The ‘X’ means that the GPU ran out of memory (OOM). The images improve in memory consumption when resized to a smaller version}
\label{fig:res_swap}
\end{figure}

The memory consumption also decreased for the compressed images (see Figures \ref{fig:res_swap}a and \ref{fig:res_swap}b). Image P1854 was able to run on both networks when compressed to 50\% or more, and only on the SSD network when compressed to 80\%. The correlation between inference time and memory consumption makes sense since images with less pixels will be quicker to analyze and require less SWAP.

\begin{figure}[t]
\centering
\begin{subfigure}{.90\columnwidth}
  \centering
  \includegraphics[width=\columnwidth]{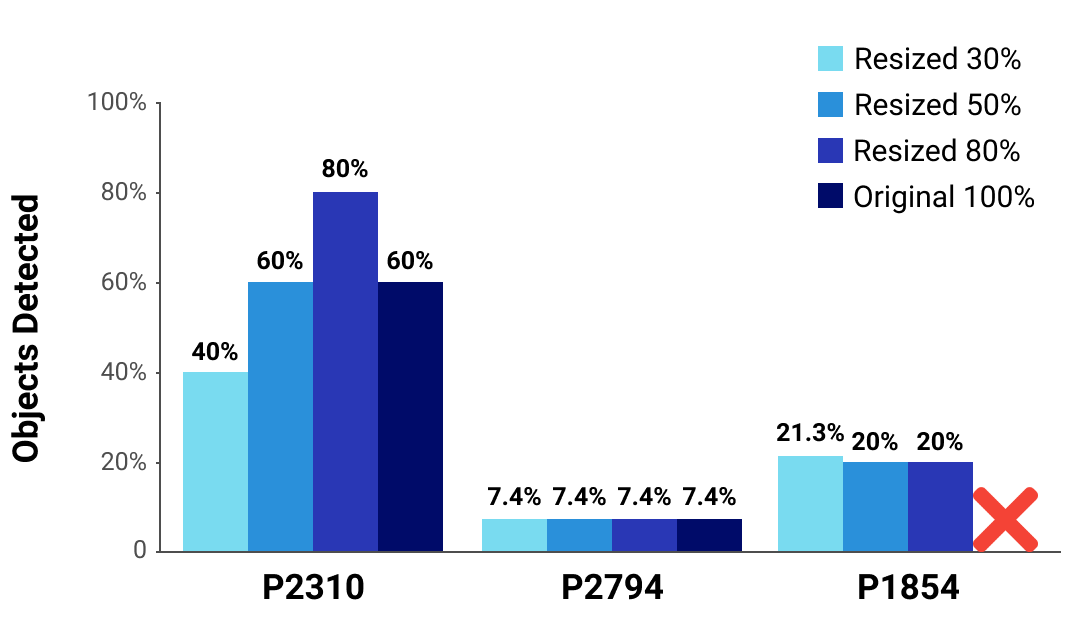}
  \subcaption{SSD}
   \label{fig:res_acc_ssd}
\end{subfigure}
\newline
\begin{subfigure}{.90\columnwidth}
  \centering
  \includegraphics[width=\columnwidth]{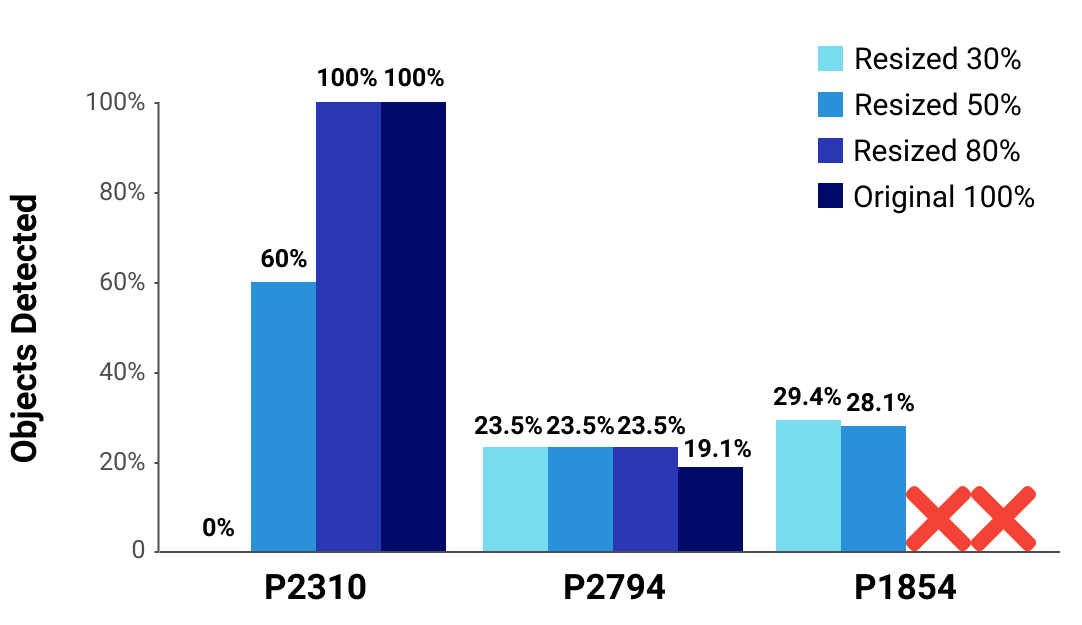}
  \subcaption{R-FCN}
  \label{fig:res_acc_rfcn}
\end{subfigure}
\caption{Objects Accurately Detected in the resized images in both models. The ‘X’ means that the GPU ran out of memory (OOM)} 
\label{fig:res_acc}
\end{figure}

However, the results in terms of accuracy were not consistent with our hypothesis in regards to the speed/ accuracy trade-off (see Figures \ref{fig:res_acc}a and \ref{fig:res_acc}b). The reason for this will be discussed in the following section. However, it was consistent with the trade-off between memory and accuracy where the images with larger pixel dimensions took up more of the SWAP memory, as expected. Furthermore, the overall time to run the full object detection code decreased at higher compression levels. For example, image P1854 on model SSD at 80\% took 49 minutes to complete the process while at 30\% it took 6 minutes. Since the aim is to place this system on a satellite that will continuously retrieve new data, low processing time per image means that more data can be processed and downlinked.

\subsection*{Aspect Ratios}
A possible reason for the surprising results in terms of accuracy for the resized images is attributed to the shape of the images. There are a lot of variations in the aspect ratios. For example, the largest image, P1854, has an aspect ratio of 3.1218 (13383:4287). Both of the neural networks are trained on square images, therefore, we want to assess the difference in performance between the original image and by fitting the image into a square. We also want to see the difference between this image in a large resolution and compressed version. This is done by splitting the original image into squares with a 10\% percent overlap to avoid any objects to be split in half which would impact the detection. Image P1854 is split into images with a ratio of 1:1 (4287:4287). The images are split using the open source DOTA Devkit\noindent{\footnote{https://github.com/CAPTAIN-WHU/DOTA\_devkit}}. By converting the image to a square shape and comparing it in a high resolution (100\%) we can see whether this affects the output.

\begin{figure}[t]
\centering
\begin{subfigure}{.95\columnwidth}
  \centering
  \includegraphics[width=\columnwidth]{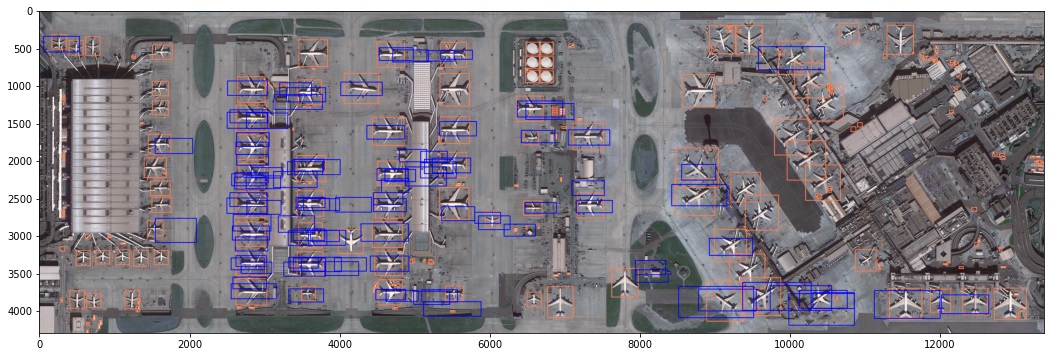}
  \subcaption{Objects detected in image P1854 original shape, 75\% resized}
  \label{fig:P1854_75}
\end{subfigure}
\newline
\begin{subfigure}{.95\columnwidth}
  \centering
  \includegraphics[width=\columnwidth]{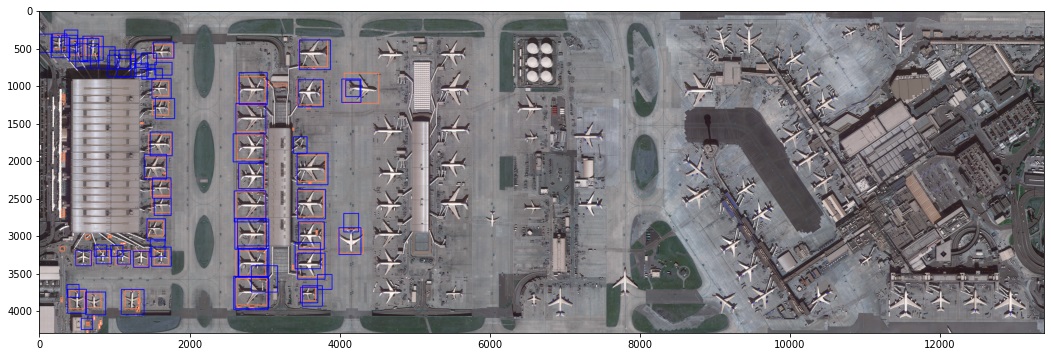}
  \subcaption{Objects detected in image P1854 squared to 4287x4287}
  \label{fig:P1854_square}
\end{subfigure}
\caption{Objects detected (in blue) and ground truth objects (in orange) in image P1854. The bounding boxes for the detected objects in image a) have a more rectangular shape than the ground truth ones, whilst in image b) they seem to match better}
\label{fig:P1854_detected}
\end{figure}

The results show that both speed and accuracy are improved in image P1854 with a square shape compared to the original rectangle shape. The improved speed and accuracy is not a surprise since the total pixel size is decreased. However, the shape of the bounding boxes give us some insight on the network. We take a closer look at P1854 80\% and P1854 with a square shape in the SSD network. P1854 80\% detected airplanes with rectangle-shaped bounding boxes while the ground truth coordinates of this image displayed squared bounding boxes (Figure \ref{fig:P1854_detected}a).

Therefore, when the threshold was set to 0.5, only one out of the 221 (4.5\%) objects were accurately detected. On the contrary, when we analyze the IoU of the squared image with the same threshold of 0.5, 32 out of 74 (43.2\%) objects were accurately detected (we include the plane in the upper-right side since the detected bounding box was cropped where the squared image ended). We can also see that the bounding boxes of the objects detected in the squared-shaped image are more similar to that of the ground truths (see Figure 20b). The same happens for R-FCN and the reason for this is that both networks have been trained on squared images so the input images in the testing are fitted into a square resolution (Mendoza and Velastin, 2018).

To explore this further, we take into account that the total pixel size of the squared image (18,378,369) is significantly smaller than that of the original image (57,372,921). Therefore, we resize the original image to a pixel size comparable with the squared image while keeping the aspect ratio the same. This resized image has the pixel size of 18,376,950 ($7575\times 2426$)\noindent{\footnote{It was not possible to get the exact same pixel size as the squared image due to the arbitrary ratio of image P1854.}}. According to the architecture of the SSD network, the size of the input should not matter as it does not require a fixed input size. However, compressing an image does reduce the quality which could impact the confidence score. We test this out by running a squared image of a lower resolution ($608\times 608$ dimension). The results show that the inference time improved by 5.7 seconds (from 6.91 seconds to 1.21 seconds), SWAP was 1.20GB smaller, and the accuracy did not change. In this way, the predictions are made on a square grid and the input images are squared as well. However, very few of the original images in the dataset are square and fitting these images into a square unfortunately takes away the uniqueness of this dataset.

\section*{CONCLUSION}

\subsection*{Summary}
In this paper we investigated two pre-trained models, SSD and R-FCN, and two compression techniques, image scaling and lossless compression. These were then applied to the validation set of DOTA (A Large-scale Dataset for Object Detection in Aerial Images). We developed a baseline by running the images on a cluster for which the performance of the Jetson Nano could be compared against. From the baseline we saw that the inference time was significantly slower when the models ran on the Jetson Nano compared to the cluster. Thus, we applied two compression techniques that modified the images in order to improve the inference time and decrease the memory consumption without a significant loss in accuracy. The accuracy was evaluated against a baseline established from running the original size of the images on the Jetson Nano and making a comparison to when the images were modified to 80\%, 50\%, and 30\% of the original image . This baseline showed that a portion of the images in the dataset were not able to run on the constrained device due to their large pixel size and the images generally had a high inference time and consumed a lot of memory. We, therefore, first applied a compression technique to resize the images in order to run the full dataset. Other images in the dataset were resized to assess their performance in terms of speed, memory consumption, and accuracy so that a speed/ memory/ accuracy trade-off could be determined. The second compression technique applied lossless compression. These images retained the total pixel size, although the file size was decreased. The results showed that by applying this compression technique, the inference time significantly decreased without any impact on accuracy. 
The first compression technique decreased the inference time for the large image P2794 by 2.77 seconds for SSD and 1.28 seconds for R-FCN, while the second compression technique decreased the inference time by 2.14 seconds for SSD and 1.16 seconds for R-FCN. This suggests that image scaling is a more effective technique for optimizing the speed. However, when taking a look at the accuracy, the lossless compression showed favorable results as the accuracy was not compromised from applying the compression technique. Therefore, an efficient way to compress the images could be to apply a combination of both compression techniques. First applying lossless compression to the image, then if inference time and memory consumption need to be decreased further, to resize the image. 

Furthermore, we developed a way to efficiently assess the accuracy using intersection over union to help determine the average precision. Results shows that 10\% of the images in the dataset are unable to run due to their large pixel dimensions. Furthermore, the processing time and memory consumption of other images in the dataset are unfavorable. By compressing the images by 59\%, we can achieve a completely runnable dataset. By testing out the resized images, we found that memory consumption and speed improved, and accuracy improved in some instances while showing questionable results in other. It was found that on average the processing time decreased by 2 seconds and 1050M less memory was used across both models after applying the image scaling technique. There was no significant decrease in accuracy. However, the overall execution time decreased by more than 40 minutes after resizing the largest images in the dataset. The R-FCN model performed better in terms of accuracy, but was slower and consumed more memory than the SSD model. Due to the architecture of the networks, we found that the accuracy of square-shaped images was higher (43.2\%) than rectangle-shaped images with large differences in width and height (4.5\%). Although accuracy was not significantly improved, applying compression techniques did indeed decrease the inference time and memory consumption, two aspects which are highly important to consider for implementing deep learning applications on constrained devices in space.

\subsection*{Future Work}
The next step of this research will evaluate more images from DOTA and other datasets. The GSD of the images could be taken into account as well as the sizes of the object instances and class categories. Future work could look into other constrained devices, compression techniques, such as lossy compression and other forms of lossless compression, and datasets. Other object detection networks, such as SSSDet (Mandal et al., 2019) or any from the Xia et al. (2018) research, could be evaluated. Furthermore, the overarching aim would be to implement the Jetson Nano on a satellite and thus working with other aspects of the device to make it space-ready would provide useful research. To do this, investigating the Orbital Edge Computing (Bradley and Brandon, 2019) implementation further or testing out the capabilities of the NVIDIA Jetson Nano or a similar device in space.

\subsection*{References}
\begin{enumerate}

\itemsep0em 
    \item Abadi, M. (2015), ‘TensorFlow: Large-Scale Machine Learning on Heterogeneous Distributed Systems’, Software available from tensorflow.org 
    \item Bawa, S. (2010), ‘Compression Using Huffman Coding’, International Journal of Computer Science and Network Security, VOL.10 No.5. 
    \item Benedek, C., Descombes, X. and Zerubia, J. (2012), ‘Building development monitoring in multitemporal remotely sensed image pairs with stochastic birth-death dynamics’, IEEE TPAMI. 34(1):33–50. 
    \item Bradley, D. and Brandon, L.(2019), ‘Orbital edge computing: Machine inference in space’, IEEE Computer Architecture Letters 18(1), 59–62. doi:10.1109/lca.2019.2907539. 
    \item Bryndin, E. (2019), ‘Robots with artificial intelligence and spectroscopic sight in hi-tech labor market’, Trends in Res 3. doi:10.15761/TR.1000149. 
    \item Bualat, M. G., Smith, T., Smith, E. E., Fong, T. and Wheeler, D. (2018), ‘Astrobee: A new tool for iss operations’, 15th Int. Conference on Space Operations. doi:10.2514/6.2018-2517. 
    \item Buchen, E. and DePasquale, D. (2014), ‘Nano/ microsatellite market assessment’, SpaceWorks Enterprises. 
    \item Campbell, J. B. and Wynne, R. H. (2011), Introduction to Remote Sensing, 5 edn, The Guilford Press. 
    \item Caulfield, B. (2014), ‘Spacex brings tegra along for the ride’, NVIDIA. 
    \item Cheng, G., Zhou, P. and Han, J. (2016), ‘Learning rotation-invariant convolutional neural networks for object detection in vhr optical remote sensing images’, IEEE Trans. Geosci. Remote Sens.. 54(12):7405–7415. 
    \item Dai, J., Li, Y., He, K. and Jian, S. (2016), ‘R-FCN: object detection via region-based fully convolutional networks’, arXiv:1605.06409v2. 
    \item Durrieu, S. and R.F., N. (2013), ‘Earth observation from space - the issue of environmental sustainability’, Elsevier Ltd. dx.doi.org/10.1016/j.spacepol.2013.07.003. 
    \item ESA (2020), ‘Robots in space’, ESA. 
    \item Evers, N. (2019), ‘Deep learning in space’, Medium. 
    \item Gibson, P. (2019), ‘Deep learning on a low power gpu’, University of Edinburgh, Project Archive. 
    \item Goodfellow, I., Bengio, Y., Courville, A. (2016), ‘Deep Learning’, The MIT Press.
    \item Greenstein, Z. (2016), ‘For astronauts, next steps on journey to space will be virtual’, NVIDIA. 
    \item Heitz, G. and Koller, D. (2008), ‘Learning spatial context: Using stuff to find things’,In ECCV pp. 30–43. 
    \item Kim, H., Park, S., Wang, J., Kim, Y. and Jeong, J.(2009), ‘Advanced bilinear image interpolation based on edge features’, IEEE. DOI: 10.1109/MMEDIA.2009.14. 
    \item Kondratyev, K., Vassilyev, O., Grigoryev, A. and Ivanian, G. (1973), ‘An analysis of the earth’s resources satellite (erts-1) data’, Remote Sensing of Environment 2, 273–283.
    \item Lam, D., Kuzma, R., McGee, K., Dooley, S., Laielli, M., Klaric, M., Bulatov, Y. and McCord, B. (2018), ‘xview: Objects in context in overhead imagery’, ArXiv. arXiv:1802.07856. 
    \item Liu, K. and Mattyus, G. (2015), ‘Fast multiclass vehicle detection on aerial images. ieee geosci. remote sensing’, Lett.. 12(9):1938–1942. 
    \item Liu, W., Anguelov, D., Erhan, D., Szegedy, C., Fu, C. and Berg, A. C. (2016), ‘SSD: single shot multibox detector’, arXiv: 1512.02325v5. 
    \item Liu, Z., Wang, H., Weng, L. and Yang, Y. (2016), ‘Ship rotated bounding box space for ship extraction from high-resolution optical satellite images with complex backgrounds’, IEEE Geosci. Remote Sensing Lett.. 13(8):1074–1078. 
    \item Loukadakis, M., Cano, J. and O’Boyle, M. (2018), ‘Accelerating Deep Neural Networks on Low Power Heterogeneous Architectures’, 11th Int. Workshop on Programmability and Architectures for Heterogeneous Multicores. 
    \item Mandal, M., Shah, M., Meena, P., Vipparthi, S. (2019), 'SSSDET: Simple Short and Shallow Network for Resource Efficient Vehicle Detection in Aerial Scenes', IEEE.
    \item Mcgovern, A. and Wagstaff, K. L. (2011), ‘Machine learning in space: Extending our reach’, Springer 84(3), 335–340. doi:10.1007/s10994-011-5249-4. 
    \item Mendoza, M. and Velastin, S. (2018), Progress in Pattern Recognition, Image Analysis, Computer Vision, and Applications, Springer, CIARP: 22nd Iberoamerican Congress. 
    \item Mundhenk, T., Konjevod, G., Sakla, W. and Boakye, K. (2016), ‘A large contextual dataset for classification, detection and counting of cars with deep learning’, In ECCV pp. 785–800. 
    \item Radu V., Kaszyk K., Wen Y., Turner J., Cano J., Crowley E. J., Franke B., O'Boyle M., and Storkey A. (2019), Performance Aware Convolutional Neural Network Channel Pruning for Embedded GPUs, IEEE International Symposium on Workload Characterization (IISWC).
    \item Razakarivony, S. and Jurie, F. (2016), ‘Vehicle detection in aerial imagery: A small target detection benchmark’, Journal of Visual Communication and Image Representation, 34:187–203. 
    \item Rotich, G., Minetto, R. and Sarkar, S. (2018), ‘Resource-constrained simultaneous detection and labeling of objects in high-resolution satellite images’, ArXiv. arXiv:1810.10110v1. 
    \item Rovder, S., Cano, J. and O’Boyle, M. (2019), ‘Optimising convolutional neural networks inference on low-powered gpus’, 12th Int. Workshop on Programmability and Architectures for Heterogeneous Multicores. 
    \item Salian, I. (2019), ‘How nasa is using gpus to visualize the mars landing: Nvidia blog’,The Official NVIDIA Blog, NVIDIA. 
    \item Sayood, K. (2006), Lossless Compression Handbook, Academic Press. 
    \item Titech (2018), ‘Real-time on-orbit image identification with deep learning (press release) (in japanese)’, The Tokyo Institute of Technology. 
    \item Tomayko, J. E. (1988), Computers in Spaceflight: the NASA Experience, NASA Technical Reports Server. 
    \item Turner J., Cano J., Radu V., Crowley E. J., O'Boyle M., and Storkey A. (2018), Characterising Across-Stack Optimisations for Deep Convolutional Neural Networks, IEEE International Symposium on Workload Characterization (IISWC).
    \item Venturini, C. C. (2017), ‘Improving mission success of cubesats’, Aerospace Report, Space System Group. 
    \item Weir, N., Lindenbaum, D., Bastidas, A., VanEtten, A., McPherson, S., Shermeyer, J., Vijay, V.K. and Tang, H. (2019), ‘Spacenet mvoi: A multi-view overhead imagery dataset’, IEEE/CVF International Conference on Computer Vision (ICCV).  
    \item Wu, S. and Li, X. (2019), ‘Iou-balanced loss functions for single-stage object detection’, arXiv:1908.05641. 
    \item Wyrwas, E. (2017), ‘Proton testing of nvidia gtx 1050 gpu’, NASA Technical Reports Server. 
    \item Xia, G.-S., Bai, X., Ding, J., Zhu, Z., Belongie, S., Luo, J., Datcu, M., Pelillo, M. and Zhang, L. (2018), ‘Dota: A large-scale dataset for object detection in aerial images’, IEEE Conference on Computer Vision and Pattern Recognition. doi:10.1109/CVPR.2018.00418.
    \item Zhao, Z.-Q., Xu, S.-t. and Wu, X. (2019), ‘Object detection with deep learning: A review’, IEEE. arXiv:1807.05511v2. 
    \item Zhu, H., Chen, X., Dai, W., Fu, K., Ye, Q. and Jiao, J. (2015), ‘Orientation robust object detection in aerial images using deep convolutional neural network’, IEEE.
\end{enumerate}

\end{document}